\documentclass{article}

\usepackage[final,nonatbib]{neurips_data_2024}

\usepackage[utf8]{inputenc} 
\usepackage[T1]{fontenc}    
\usepackage{hyperref}       
\usepackage{url}            
\usepackage{booktabs}       
\usepackage{amsfonts}       
\usepackage{nicefrac}       
\usepackage{microtype}      
\usepackage[table]{xcolor}         

\usepackage{booktabs}
\usepackage{caption}
\usepackage{subcaption}
\usepackage{wrapfig}
\usepackage{etoolbox}
\usepackage{fancyvrb}
\usepackage[inline]{enumitem}
\usepackage{amsmath}
\usepackage{graphicx}
\usepackage{tabularray} 
\usepackage{multirow}
\usepackage{makecell}
\usepackage{diagbox}
\usepackage{mdframed}

\usepackage{arydshln}

\usepackage{colortbl}

\usepackage{environ}
\usepackage[ruled,vlined]{algorithm2e}
\usepackage{algcompatible}
\usepackage[noend]{algpseudocode}

\usepackage{xspace}
\usepackage{mathtools}

\usepackage{blindtext}

\usepackage{pifont} 
\newcommand{\notick}{\ding{55}}
\newcommand{\yestick}{\ding{51}}
\definecolor{urlcolor}{rgb}{0.0, 0.53, 0.74}

\NewEnviron{boxcode2col}[4]{
    \begin{center}
        \scalebox{#3}{
            \setlength\tabcolsep{4pt}
            \renewcommand{\arraystretch}{#4}
            \begin{tabular}{|p{#1}p{#2}|}   
                \hline
                \BODY
                [1ex]
                \hline
            \end{tabular}
        }
    \end{center}
}

\NewEnviron{boxcode}[3]{
    \begin{center}
        \scalebox{#2}{
            \setlength\tabcolsep{4pt}
            \renewcommand{\arraystretch}{#3}
            \begin{tabular}{!{\vrule}p{#1}!{\vrule}}
                \hline
                \vspace{0.1mm}    
                \BODY
                \\\hline
    		\end{tabular}
    	}
    \end{center}
}

\newcommand{\textcode}[1]{{\fontfamily{cmtt}\selectfont #1}\xspace}

\definecolor{tabblue}{HTML}{1F77B4}

\newcommand{\TechGPTFour}{\ensuremath{\textsc{GPT4}_{\text{text}}}}
\newcommand{\TechGPTFourV}{\ensuremath{\textsc{GPT4}_{\text{vis}}}}
\newcommand{\TechGPTFourVCombined}{\ensuremath{\textsc{GPT4}_{\text{vis+text}}}}
\newcommand{\TechChatGPT}{\ensuremath{\textsc{GPT3.5}}}
\newcommand{\TechLlamaSeven}{\ensuremath{\textsc{CodeLlama-7B-Instruct}}}

\newcommand{\TechGPTFourO}{\ensuremath{\textsc{GPT4O}_{\text{text}}}}
\newcommand{\TechGPTFourOVision}{\ensuremath{\textsc{GPT4O}_{\text{vis}}}}
\newcommand{\TechGPTFourOCombined}{\ensuremath{\textsc{GPT4O}_{\text{vis+text}}}}

\newcommand{\TechLlamaThreeEight}{\ensuremath{\textsc{Llama3-8B-Instruct}}}
\newcommand{\TechLlava}{\ensuremath{\textsc{Llava1.5-7B}}}

\newcommand{\TechHocFinetuneNoExpThree}{\ensuremath{\textsc{LlamaCT:HoC}}}
\newcommand{\TechHocFinetuneExpThree}{\ensuremath{\textsc{LlamaCT:HoC}_{\text{exp}}}}
\newcommand{\TechPlainFinetuneThree}{\ensuremath{\textsc{LlamaCT:HoC+MCQ}}}
\newcommand{\TechTopFinetuneThree}{\ensuremath{\textsc{LlamaCT:HoC+MCQ}_{\text{exp}}}}
\newcommand{\TechHierFinetuneThree}{\ensuremath{\textsc{LlamaCT:HoC+MCQ+Aug}_{\text{exp}}}}
\newcommand{\TechHierEmuFinetuneEmuInfThree}{\ensuremath{\textsc{LlamaCT:HoC+MCQ+Aug}_{\text{exp*}}}}

\newcommand{\TechLlamaCT}{\ensuremath{\textsc{LlamaCT}}}

\newcommand{\TechHuman}{\ensuremath{\textsc{Grade7}_{\text{top25}}}}
\newcommand{\TechHumanAll}{\ensuremath{\textsc{GradeAll}}}
\newcommand{\TechChance}{\ensuremath{\textsc{Naive}}}

\newcommand{\grid}{\ensuremath{\text{\textcode{G}}}}
\newcommand{\gridSet}{\ensuremath{\text{\{\textcode{G}{}\}}}}

\newcommand{\code}{\ensuremath{\text{\textcode{C}}}}
\newcommand{\codeBlocks}{\ensuremath{\text{\textcode{C}}_{\text{blocks}}}}

\newcommand{\codeSize}{\ensuremath{\text{\textcode{C}}_{\text{size}}}}
\newcommand{\codeSketch}{\ensuremath{\text{\textcode{C}}_{\text{sketch}}}}

\newcommand{\limSize}{\ensuremath{\text{\textcode{maxSize}}}}
\newcommand{\codeStore}{\ensuremath{\text{\textcode{Store}}}}

\newcommand{\codespace}{\ensuremath{\mathbb{C}}}

\newcommand{\codeGridSetPairs}{\ensuremath{\text{(\textcode{C},\{\textcode{G}{}\})}}}
\newcommand{\codeGridPair}{\ensuremath{\text{(\textcode{C},\textcode{G}{})}}}

\newcommand{\DSLWhile}{\textcode{\textsc{While}}}

\newcommand{\DSLMove}{\textcode{move}}
\newcommand{\DSLTurnLeft}{\textcode{turnLeft}}
\newcommand{\DSLTurnRight}{\textcode{turnRight}}
\newcommand{\DSLRepeat}{\textcode{\textsc{Repeat}}}
\newcommand{\DSLRepeatUntil}{\textcode{\textsc{RepeatUntil}}}
\newcommand{\DSLIf}{\textcode{\textsc{If}}}
\newcommand{\DSLIfElse}{\textcode{\textsc{IfElse}}}

\newcommand{\hocTypeBold}{\textbf{\textsc{HoC}}}
\newcommand{\hocType}{\textsc{HoC}}

\newcommand{\aceTypeBold}{\textbf{\textsc{ACE}}}
\newcommand{\aceType}{\textsc{ACE}}

\newcommand{\ctTypeBold}{\textbf{\textsc{CT-Test}}}
\newcommand{\ctType}{\textsc{CT-Test}}

\newcommand{\Analyzing}{\emph{Analyzing}}
\newcommand{\Evaluating}{\emph{Evaluating}}
\newcommand{\Creating}{\emph{Creating}}

\newcommand{\aceTypeA}{\textsc{ACE}[01-07]}
\newcommand{\aceTypeE}{\textsc{ACE}[08-14]}
\newcommand{\aceTypeC}{\textsc{ACE}[15-21]}

\newcommand{\descr}{{\text{\textcode{D}}}}
\newcommand{\answer}{{\text{\textcode{Ans}}}}
\newcommand{\distractors}{{\text{\textcode{\{Dis\}}}}}

\definecolor{TutorColour}{RGB}{105, 105, 105}
\definecolor{PCFGColour}{RGB}{204, 153, 24}

\definecolor{RoyalPurple}{RGB}{120, 81, 169}
\definecolor{OliveGreen}{RGB}{128, 128, 0}

\newcommand{\target}{{\text{\ensuremath{\text{\textcode{Task}}_{\text{out}}}}}}
\newcommand{\inp}{{\text{\ensuremath{\text{\textcode{Task}}_{\text{in}}}}}}

\title{Benchmarking Generative Models on Computational Thinking Tests in Elementary Visual Programming}

\author{
        Victor-Alexandru P{\u a}durean \\
        MPI-SWS\\
        \texttt{vpadurea@mpi-sws.org} \\
        \And
        Adish Singla \\
        MPI-SWS \\
        \texttt{adishs@mpi-sws.org} \\
      }

\begin{document}

\maketitle

\newtoggle{MainSuppContent}
\settoggle{MainSuppContent}{true}


\iftoggle{MainSuppContent}{
\begin{abstract}
Generative models have demonstrated human-level proficiency in various benchmarks across domains like programming, natural sciences, and general knowledge. Despite these promising results on competitive benchmarks, they still struggle with seemingly simple problem-solving tasks typically carried out by elementary-level students. How do state-of-the-art models perform on standardized programming-related tests designed to assess computational thinking and problem-solving skills at schools? In this paper, we curate a novel benchmark involving computational thinking tests grounded in elementary visual programming domains. Our initial results show that state-of-the-art models like GPT-4o and Llama3 barely match the performance of an average school student. To further boost the performance of these models, we fine-tune them using a novel synthetic data generation methodology. The key idea is to develop a comprehensive dataset using symbolic methods that capture different skill levels, ranging from recognition of visual elements to multi-choice quizzes to synthesis-style tasks. We showcase how various aspects of symbolic information in synthetic data help improve fine-tuned models' performance. We will release the full implementation and datasets to facilitate further research on enhancing computational thinking in generative models.
\end{abstract}

\section{Introduction} \label{sec.introduction}
The recent advances in generative models and large language models (LLMs) have the potential to positively impact a wide variety of domains, such as medicine \cite{DBLP:journals/corr/abs-2212-13138,DBLP:conf/nips/LiWZULYNPG23,DBLP:journals/bib/LuoSXQZPL22}, arts \cite{DBLP:journals/corr/abs-2402-09750,DBLP:conf/uist/AngertSHPS23}, and education \cite{khanmigo,DBLP:journals/corr/abs-2310-10690,DBLP:journals/corr/abs-2402-01580,DBLP:conf/lak/PhungPS0CGSS24}. This potential is reflected by their success on a wide range of popular competitive benchmarks assessing their knowledge of natural sciences and day-to-day facts \cite{GPT4o,Llama3,DBLP:conf/iclr/HendrycksBBZMSS21,DBLP:conf/acl/ZellersHBFC19,DBLP:journals/corr/abs-1803-05457} and their skills in programming. For example, GPT-4o \cite{GPT4o} is capable of obtaining a high accuracy on two popular programming benchmarks: $90.2\%$ on HumanEval \cite{DBLP:journals/corr/abs-2107-03374} and $87.5\%$ on MBPP \cite{DBLP:journals/corr/abs-2108-07732}. Previous studies also showed that GPT-4 \cite{GPT4} is capable of passing assessments in higher education programming courses, achieving course totals greater than $79\%$ \cite{DBLP:conf/icer/SavelkaAABS22}.

Despite these promising results, state-of-the-art models struggle with seemingly simple tasks. These models often underperform in tasks requiring mathematical reasoning, planning, and problem-solving \cite{DBLP:journals/corr/abs-2303-12712,DBLP:conf/nips/HendrycksBKABTS21,DBLP:journals/corr/abs-2305-13160,DBLP:conf/nips/ValmeekamMHSK23}. For example, they fail to solve planning tasks involving stacking of colored blocks \cite{DBLP:conf/nips/ValmeekamMSK23}. Moreover, generative models often face problems with basic algebra and counting \cite{DBLP:journals/corr/abs-2303-12712}, or coming up with correct codes in visual programming domains \cite{DBLP:journals/corr/abs-2308-02522}, tasks which can successfully be carried out by elementary-level school students. These weaknesses seem to contradict the generative models' impressive performance in complex programming tasks. Based on these observations, we aim to study how generative models tackle programming tasks specifically designed to foster computational thinking and problem-solving skills in elementary-level students. This leads to our main research question: \emph{How do state-of-the-art models perform on standardized programming-related tests designed to assess computational thinking and problem-solving skills at schools?}

In this paper, we introduce a novel benchmark for assessing generative models' computational thinking and problem-solving capabilities. We conduct extensive experiments with various models and our results show that state-of-the-art models struggle with the computational thinking tests in our benchmark. Figure~\ref{fig.illustration.radar} illustrates how GPT-4o barely matches the performance of an average school student, with Llama3 \cite{Llama3} performing even worse.

\looseness-1We make the following contributions towards improving the models' performance on computational thinking tests:
\begin{enumerate*}[label={(\arabic*)},leftmargin=*]

\item We introduce a novel data generation methodology based on symbolic methods. An important aspect of the generated dataset is that it captures different skill levels, ranging from recognition of visual elements to multi-choice questions to synthesis-style tasks.

\item We fine-tune Llama3-8B and obtain the \TechLlamaCT{} family of models, the best of which achieves an accuracy on par with GPT-4o (see Figure~\ref{fig.illustration.radar}). We further analyze how various aspects of symbolic information in our synthetic dataset help improve the fine-tuned models' performance.

\item We will release the data and implementation to promote further research on enhancing computational thinking in generative models.\footnote{GitHub repo: \url{https://github.com/machine-teaching-group/neurips2024-benchmark-ct}.}

\end{enumerate*}

\begin{figure*}
    \begin{subfigure}{0.49\textwidth}
        \centering
        \includegraphics[width=0.79\textwidth, trim={0.46cm 0.93cm 0.08cm 0.1cm},clip]{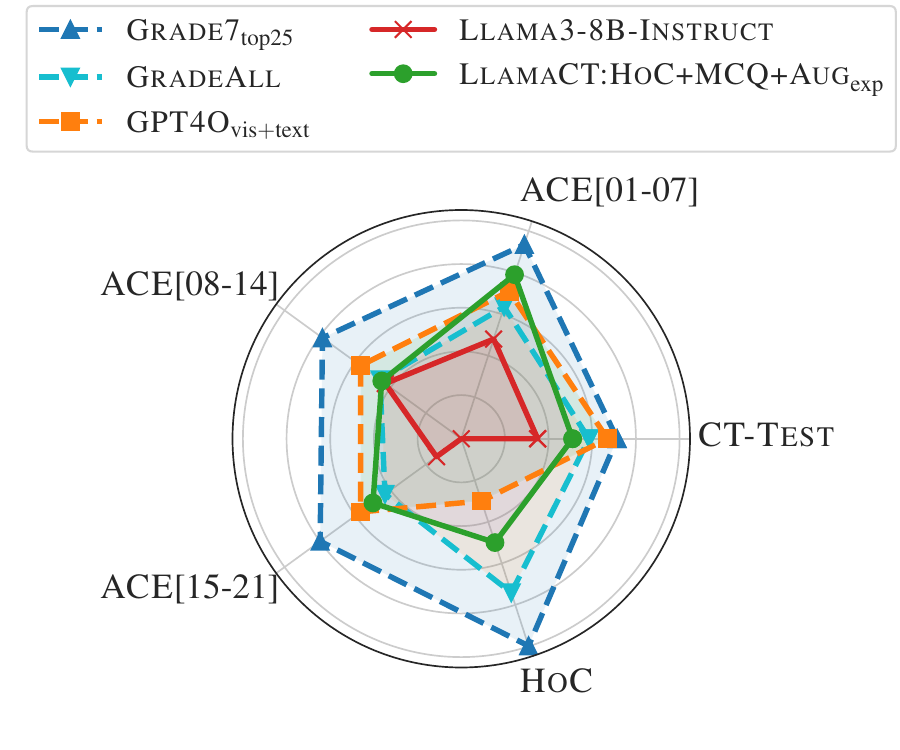}
        \vspace{-1.5mm}
        \caption{Performance on computational thinking tests.}
        \vspace{1mm}
        \label{fig.illustration.radar}
        \end{subfigure}
    %
    \begin{subfigure}{0.49\textwidth}
        \centering
        \setlength{\fboxsep}{0.05pt}\fbox{\includegraphics[width=0.95\textwidth]{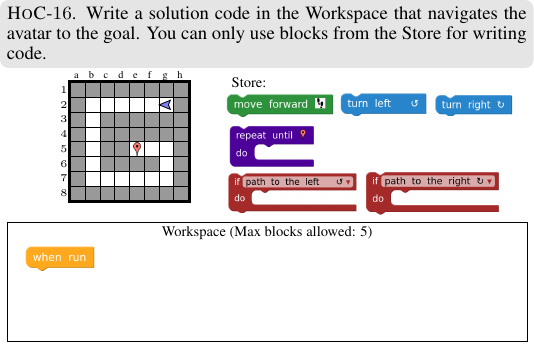}}
        \vspace{-1mm}
        \caption{A problem-solving task from \hocType{}.}
        \vspace{1mm}
        \label{fig.illustration.hoc}
    \end{subfigure}
    \\
    \begin{subfigure}{0.49\textwidth}
        \centering
        \setlength{\fboxsep}{0.05pt}\fbox{\includegraphics[width=0.95\textwidth]{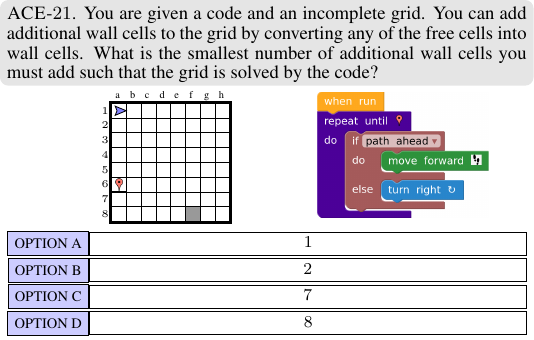}}
        \vspace{-1mm}
        \caption{A multi-choice question task from \aceType{}.}
        \label{fig.illustration.ace}
    \end{subfigure}
    %
    \begin{subfigure}{0.49\textwidth}
        \centering
        \setlength{\fboxsep}{0.05pt}\fbox{\includegraphics[width=0.95\textwidth]{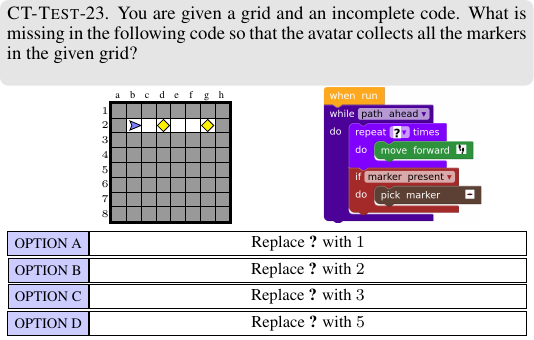}}
        \vspace{-1mm}
        \caption{A multi-choice question task from \ctType{}.}
        \label{fig.illustration.ct}
    \end{subfigure}
    \caption{\textbf{(a)} shows the performance of school students compared to various models on a scale of $0$ to $100$. We break down the \aceType{} \cite{DBLP:conf/sigcse/GhoshMS24} test into its constituent parts: \Analyzing{} (\aceTypeA{}), \Evaluating{} (\aceTypeE{}), and \Creating{} (\aceTypeC{}). \textbf{(b)} shows \textsc{Maze16} from \emph{Hour of Code:Maze Challenge} (\hocType{})~\cite{hourofcode_maze,codeorg}, an example of a solution synthesis task. \textbf{(c)} shows a \Creating{} multi-choice question task from the \aceType{} test. \textbf{(d)} shows a multi-choice question task from the \ctType{}~\cite{gonzalez2015computational,DBLP:journals/chb/Roman-GonzalezP17}.}
    \label{fig.illustration}
    \vspace{-2mm}    
\end{figure*}

\section{Related Work} \label{sec.relatedwork}


\begin{figure*}
\centering{
\scalebox{0.8}{
\setlength\tabcolsep{4pt}
\renewcommand{\arraystretch}{1.2}  
    \begin{tabular}{l c c c r c c }
    \toprule 
    \vspace{-1.50mm}
    \textbf{Work} & \textbf{Domain} & \textbf{Evaluation tasks} & \textbf{Multimodal} & \multicolumn{1}{c}{\textbf{Benchmark}} & \textbf{Trained} & \textbf{Human} \\
    & & & & \multicolumn{1}{c}{\textbf{evaluation size}} & \textbf{model} & \textbf{comparison}
    \\
    \midrule
    {\cellcolor{green!60!black!15}}{Our Work} & {\cellcolor{green!60!black!15}}{visual} & {\cellcolor{green!60!black!15}}{code synthesis,} & {\cellcolor{green!60!black!15}}{\yestick{}} & 
    {\cellcolor{green!60!black!15}}{$65$ \ \ \ \ } & {\cellcolor{green!60!black!15}}{\yestick{}} & {\cellcolor{green!60!black!15}}{\yestick{}} \\
    {\cellcolor{green!60!black!15}}{} & {\cellcolor{green!60!black!15}}{programming} & {\cellcolor{green!60!black!15}}{multiple choice questions}  & {\cellcolor{green!60!black!15}}{} & {\cellcolor{green!60!black!15}}{} & {\cellcolor{green!60!black!15}}{} & {\cellcolor{green!60!black!15}}{} \\
    \hline
    HumanEval~\cite{DBLP:journals/corr/abs-2107-03374} & programming & Python code writing & \notick{} & $164$ \ \ \ \  & \yestick{} & \notick{} \\
    \hline
    MMCode~\cite{li2024mmcode} & programming & Python code writing & \yestick{} & $3,548$ \ \ \ \  & \notick{} & \notick{} \\
    \hline
    MathVista~\cite{DBLP:journals/corr/abs-2310-02255} & mathematics & multiple choice questions, & \yestick{} & $6,141$ \ \ \ \  & \notick{} & \yestick{} \\
    & & integer free form questions & & & & \\
    \hline
    MMMU~\cite{DBLP:journals/corr/abs-2311-16502} & diverse & multiple choice questions, & \yestick{} & $11,500$ \ \ \ \  & \notick{} & \yestick{} \\
    & & open answer questions & & & & \\
    \hline
    HoC+Karel~\cite{DBLP:journals/corr/abs-2308-02522} & visual &  code synthesis, & \notick{} & $30$ \ \ \ \  & \notick{} & \notick{} \\
    & programming & tracing, grid synthesis & & & & \\
    \bottomrule
    \end{tabular}
    } 
    }
    \caption{
    Comparison of our work with related benchmarks. The first column shows the name of the benchmark and the work it was introduced in. ``Domain'' specifies the domain for which the benchmark was designed, and ``Evaluation tasks'' outlines the tasks involved. ``Multimodal'' indicates if the benchmark includes both visual and textual data. ``Benchmark evaluation size'' shows the number of samples in each benchmark. ``Trained model'' notes whether any model is trained in the work, and ``Human comparison'' indicates if model performance was compared to that of humans.
    }
    \label{fig.relatedwork}
    \vspace{-3mm}    
\end{figure*}

We identify two key research themes in the literature: one focuses on the programming capabilities of generative models, and the other focuses on their general reasoning capabilities. To our knowledge, this paper is the first to evaluate numerous generative models on a comprehensive set of computational thinking tests grounded in elementary visual programming. Figure~\ref{fig.relatedwork} presents a comparison between our work and the most relevant benchmarks in the literature.

\looseness-1\textbf{Benchmarks assessing programming capabilities.} Several works benchmark the programming capabilities of models, with popular examples of benchmarks including HumanEval~\cite{DBLP:journals/corr/abs-2107-03374}, MBPP~\cite{DBLP:journals/corr/abs-2108-07732}, and APPS~\cite{DBLP:conf/nips/HendrycksBKMAGB21}. For example, HumanEval~\cite{DBLP:journals/corr/abs-2107-03374} focuses on Python code generation but lacks multimodal elements and human comparisons, as shown in Figure~\ref{fig.relatedwork}. A more recent benchmark, MMCode~\cite{li2024mmcode}, includes visual information in traditional coding tasks to assess multimodal model capabilities. However, this benchmark does not include comparisons between model and human performance, nor does it explore potential improvements in model performance through fine-tuning or other methods. Besides program generation, other benchmarks handle code completion, translation, summarization, debugging or explanation generation \cite{DBLP:journals/corr/abs-2308-07124,DBLP:journals/corr/abs-2306-17156,DBLP:journals/corr/abs-2401-04621,DBLP:journals/corr/abs-2306-03091}, thus analyzing numerous programming-related tasks. However, these works typically focus on generating code or explanations and do not evaluate the core computational thinking and problem-solving skills of models. In contrast, our paper seeks to provide a deeper understanding of these capabilities. Additionally, we train models on our dataset and compare their performance to human counterparts, addressing gaps in previous studies.

\textbf{Benchmarks assessing reasoning capabilities.} For general reasoning, benchmarks like MathVista~\cite{DBLP:journals/corr/abs-2310-02255} and MMMU~\cite{DBLP:journals/corr/abs-2311-16502} assess models on tasks involving multiple-choice and free-form questions, with a focus on multimodal data. These, along with other benchmarks, evaluate reasoning in fields such as mathematics and the natural sciences \cite{DBLP:journals/corr/abs-1803-05457,DBLP:conf/nips/HendrycksBKABTS21,DBLP:journals/corr/abs-2311-16502,DBLP:journals/corr/abs-2310-02255}, planning \cite{DBLP:conf/nips/ValmeekamMHSK23,DBLP:conf/nips/ValmeekamMSK23,DBLP:conf/icml/HuangAPM22}, and causal reasoning \cite{DBLP:conf/nips/JinCLGKLBAKSS23,jin2024can}. Our benchmark goes beyond these by including a variety of tasks that assess computational thinking through visual programming. This relatively underexplored area can offer intriguing insights into the reasoning capabilities of generative models. Previous efforts~\cite{DBLP:journals/corr/abs-2308-02522} address visual tasks without the multimodal component and do not include human comparisons. In contrast, our benchmark integrates multimodal tasks combining both programming and visual reasoning, while also comparing models directly against human performance, offering a more comprehensive evaluation of reasoning abilities in generative models.

\section{Computational Thinking Tests in Elementary Visual Programming} \label{sec.background}

This section first provides a background on visual programming, and then introduces the sources we use for curating our benchmark and as the basis for our synthetic dataset generation methodology.

\subsection{Preliminaries and Definitions for Elementary Visual Programming}

\textbf{The space of grids.}
A visual grid, denoted as \grid{}, includes an avatar with an initial position (row, column) and orientation (north, east, south, west), alongside free cells, wall cells, and a goal or multiple markers. The avatar is required to reach the goal or interact with the markers. The resulting grid space includes visual grids based on \hocType{} by Code.org~\cite{hourofcode_maze,codeorg}, such as the grids in Figures~\ref{fig.illustration.hoc}~and~\ref{fig.illustration.ace}, and Karel~\cite{pattis}, such as the grid in Figure~\ref{fig.illustration.ct}.

\textbf{The space of codes.}
The set of valid codes \codespace{} is defined via a domain-specific language (DSL).  We adopt DSLs previously used in literature for visual programming \cite{padurean2024neural,DBLP:conf/iclr/BunelHDSK18,DBLP:conf/nips/AhmedCEFGRS20}. A code $\code \in \codespace$ is characterized by its size $\codeSize$, utilized constructs $\codeBlocks$, and programming concepts exercised in terms of its nesting structure $\codeSketch$. For example, the code in Figure~\ref{fig.illustration.ace} uses $\codeSize = 5$ blocks, with constructs $\codeBlocks = \{\text{\DSLMove, \DSLTurnRight, \DSLRepeatUntil, \DSLIfElse}\}$, and is structured as $\codeSketch=\DSLRepeatUntil{}\textcode{\{\DSLIfElse{}\}}$. Executing a code on a grid generates a sequence of avatar locations, referred to as trace, along with a sequence of basic actions executed i.e., constructs from $\{\text{\DSLMove, \DSLTurnLeft, \DSLTurnRight}\}$. A code is considered to solve a grid if it successfully navigates the avatar to the goal or interacts correctly with the markers (e.g., collects them when intended).

\textbf{Solution synthesis tasks.}
\looseness-1A solution synthesis task is defined by the following elements: a grid \grid{}, an allowed set of constructs called \codeStore{}, and a maximum code size \limSize{}. The objective is to write a solution code \code{} that successfully solves \grid{} while respecting $\codeBlocks \subseteq \codeStore$ and $\codeSize \le \limSize$. Figure~\ref{fig.illustration.hoc} exemplifies a solution synthesis task, where a solution code \code{} should solve \grid{}, have $\codeBlocks \subseteq \{\text{\DSLMove, \DSLTurnRight, \DSLTurnLeft, \DSLRepeatUntil, \DSLIfElse}\}$ and $\codeSize \leq 5$.

\textbf{Multi-choice question tasks.}
A multi-choice question (MCQ) task is defined by the following elements: a text description, a set of grids or codes, one correct option, and three distractor options. The objective is to choose the correct option out of four options. For example, Figures~\ref{fig.illustration.ace}~and~\ref{fig.illustration.ct} have a text description inside the gray area, a given grid and a given code, and four options. The correct option for Figure~\ref{fig.illustration.ace} is Option A, and the correct option for Figures~\ref{fig.illustration.ct} is Option A as well.

\subsection{Three Different Computational Thinking Tests}
\label{sec.background.tests}

\looseness-1Our benchmark is based on two pedagogically validated computational thinking tests comprising multiple-choice question tasks \cite{DBLP:conf/sigcse/GhoshMS24,gonzalez2015computational,DBLP:journals/chb/Roman-GonzalezP17} and a popular curriculum comprising code-writing tasks \cite{hourofcode_maze}. Henceforth, we refer to these three as tests, and we will use them throughout the paper to measure the performance of generative models. These tests have been carefully designed by educational experts to assess or teach a diverse set of skills in elementary visual programming within the duration of a typical one-hour school lesson. They are representative of computational thinking in this domain, providing valuable data on student performance, which we can use as a basis to benchmark the performance of generative models.
Figure~\ref{fig.background.distribution_table} gives an overview of the programming concepts \codeSketch{} utilized by tasks in each test. Next, we provide details for each test.

\textbf{HoC.}
This test includes 20 code-writing tasks from Code.org's popular block-based visual programming lesson \emph{Hour of Code:Maze Challenge}~\cite{hourofcode_maze}. The tasks mainly cover concepts such as basic actions, \DSLRepeat{} and \DSLRepeatUntil{} loops, as well as \DSLIf{} and \DSLIfElse{} branching (see Figure~\ref{fig.background.distribution_table}). This curriculum has been used by millions of learners to get acquainted with programming and to assess students' programming background \cite{DBLP:conf/sigcse/GhoshMS24,hourofcode_maze,codeorg}.

\begin{figure*}[t!]
\centering
    \scalebox{0.75}{
        \setlength\tabcolsep{10pt}
        \renewcommand{\arraystretch}{0.9}
        \begin{tabular}{lrrrrr}
            \toprule
            
            \multirow[c]{1}{*}[0mm]{{\textbf{Concepts}}} &            
            \multicolumn{1}{c}{\hocTypeBold{}} &
            \multicolumn{3}{c}{\aceTypeBold{}} &
            \multicolumn{1}{c}{\ctTypeBold{}} \\

            \cmidrule(lr){3-5}{}
            
            &
            &
            \Analyzing{} &
            \Evaluating{} &
            \Creating{} &
            \\
            
            &
            &
            \aceTypeA{} &
            \aceTypeE{} &
            \aceTypeC{} &
            \\
            
            \midrule 

            Basic actions &
            H01--H05&
            Q01&
            Q08&
            Q15&
            P01--P04\\

            \midrule

            \DSLRepeat{}\textcode{\{\}} &
            H06--H09&
            Q02, Q05&
            Q12&
            Q16&
            P05, P06\\

            \DSLRepeatUntil{}\textcode{\{\}} &
            H10--H13&
            Q06&
            Q09&
            Q17, Q18&
            P09, P10\\

            \midrule

            \DSLRepeatUntil{}\textcode{\{\DSLIf{}\}} &
            H14--H17&
            Q07&
            Q10&
            Q19&
            P13, P14\\

            \DSLRepeatUntil{}\textcode{\{\DSLIfElse{}\}} &
            H18, H19&
            Q04&
            Q11, Q14&
            Q20, Q21&
            P17, P18\\

            \DSLRepeatUntil{}\textcode{\{\DSLIfElse{}\textcode{\{\DSLIfElse{}\}}{}\}} &
            H20&
            &
            &
            &
            P19, P20\\

            \midrule

            \DSLRepeat{}\textcode{\{\DSLRepeat{}\}} &
            &
            &
            Q13&
            &
            P08, P27, P28
            \\

            \DSLRepeat{}\textcode{\{\DSLIf{}\}} &
            &
            Q03&
            &
            &
            \\

            \midrule
            
            \DSLRepeatUntil{}\textcode{\{\DSLIf{}\textcode{; }\DSLIf{}\}} &
            &
            &
            &
            &
            P16\\

            \DSLRepeatUntil{}\textcode{\{\DSLRepeat{}\}} &
            &
            &
            &
            &
            P11\\

            \DSLRepeatUntil{}\textcode{\{\DSLIf{}\textcode{\{\DSLRepeat{}\}}{}\}} &
            &
            &
            &
            &
            P15\\

            \DSLWhile{}\textcode{\{\}; }\DSLRepeat{}\textcode{\{\}} &
            &
            &
            &
            &
            P21\\

            \DSLWhile{}\textcode{\{\DSLRepeat{}\textcode{; }\DSLRepeat{}\}} &
            &
            &
            &
            &
            P22\\

            \DSLWhile{}\textcode{\{\DSLRepeat{}\textcode{; }\DSLIf{}\}} &
            &
            &
            &
            &
            P23\\

            \DSLWhile{}\textcode{\{\DSLIf{}\textcode{\{\DSLWhile{}\}}{}\}} &
            &
            &
            &
            &
            P24\\

            \bottomrule   
        \end{tabular}
        }
    \caption{\looseness-1Programming concepts $\codeSketch$ required for solving tasks in \hocType{}~\cite{hourofcode_maze}, \aceType{}~\cite{DBLP:conf/sigcse/GhoshMS24}, and \ctType{}~\cite{gonzalez2015computational,DBLP:journals/chb/Roman-GonzalezP17}. \hocType{} comprises code-writing tasks. \aceType{} and \ctType{} comprise multi-choice question tasks. \aceType{} is further split according to the higher cognitive levels of Bloom's taxonomy \cite{bloom1956taxonomy,krathwohl2002revision}. }
    \label{fig.background.distribution_table}    
    \vspace{-2mm}
\end{figure*}

\textbf{ACE.}
\looseness-1This test includes 21 multi-choice question tasks from the \aceType{} test, which was designed to evaluate higher cognitive levels of Bloom's taxonomy: \Analyzing{}, \Evaluating{}, and \Creating{} \cite{DBLP:conf/sigcse/GhoshMS24, bloom1956taxonomy,krathwohl2002revision}. These tasks were selected from a larger pool to ensure balanced coverage of cognitive levels and programming concepts, being validated using standardized pedagogical tools. Figure~\ref{fig.background.distribution_table} categorizes each task by cognitive level and programming concepts covered. 

\textbf{CT-Test.}
\looseness-1This test is based on \ctType{}, one of the earliest and most popular computational thinking tests in block-based visual programming \cite{gonzalez2015computational,DBLP:journals/chb/Roman-GonzalezP17}. Out of 28 tasks in the original set, we curate 24 tasks compatible with our definitions and representation. Figure~\ref{fig.background.distribution_table} shows the programming concepts covered, with the original task numbering: if its number is not in the table, we have not included the task.
\section{\looseness-1Synthetic Data Generation to Fine-tune Models for Computational Thinking} \label{sec.dataset}

In this section, we introduce our novel data generation methodology for computational thinking and problem-solving skills. With the resulting data (see Figure~\ref{fig.data.distribution}), we aim to fine-tune models to increase performance on all three tests.
Next, we present our three main methods for generating data.

\subsection{Synthetic Data for Solution Synthesis}
\label{sec.dataset.solsyn}

We first generate data for solution synthesis tasks. Our process will start with generating a dataset of pairs \codeGridSetPairs{}, where \code{} is a code and \gridSet{} is a set of grids solved by \code{}. To obtain \codeGridSetPairs{}, we employ existing techniques for synthesizing code \code{} and grid \grid{} \cite{padurean2024neural,DBLP:conf/iclr/BunelHDSK18,DBLP:conf/nips/AhmedCEFGRS20}. We then split the sets into pairs of one solution code and one grid \codeGridPair{}. We extract \codeStore{} and \limSize{} from code \code{}. Then, we treat $(\grid, \codeStore, \limSize)$ as input for the task, and keep \code{} as target output.

To enhance the fine-tuning process, 
we aim to train the model to first produce a trace and sequence of basic actions that the avatar should execute to reach the goal, and then to produce the solution code. We refer to the trace and sequence of basic actions as an explanation for the produced answer. This method is grounded in previous research, which has shown that smaller models benefit from richer signals while being fine-tuned, leading to more careful reasoning at inference \cite{DBLP:journals/corr/abs-2306-02707,DBLP:journals/corr/abs-2311-11045}. However, unlike literature, we cannot rely on more powerful models like GPT-4 to produce these explanations, as state-of-the-art models struggle with computational thinking (see Figure~\ref{fig.illustration.radar}). So, we rely on symbolic methods such as executing codes on grids via an emulator to produce correct traces and basic action sequences as explanations.

\begin{figure*}[t!]
\centering
    \begin{subfigure}[b]{.5\textwidth}
	\centering
        \centering
    \scalebox{0.74}{
        \setlength\tabcolsep{4pt}
        \renewcommand{\arraystretch}{0.85}
        \begin{tabular}{lrrr}    
            \toprule
            \multicolumn{1}{c}{Synthetic data} & Original & Selected & Percentage \\
            & Size & Size & \\
            \midrule
            Solution synthesis &
            $7,576$ &
            $7,576$ &
            $6.77\%$ \\
            \midrule
            
            Multi-choice questions (MCQ) &
            $9,223$ &
            $9,223$ &
            $8.25\%$ \\
            
            \color{black!65} \Analyzing{} MCQ (A) &
            \color{black!65} $2,779$ &
            \color{black!65} $2,779$ &
            \color{black!65} $2.49\%$ \\
            
            \color{black!65} \Evaluating{} MCQ (E) &
            \color{black!65} $2,072$ &
            \color{black!65} $2,072$ &
            \color{black!65} $1.85\%$ \\
            
            \color{black!65} \Creating{} MCQ (C) &
            \color{black!65} $4,372$ &
            \color{black!65} $4,372$ &
            \color{black!65} $3.91\%$ \\
            \midrule
            
            Fine-grained: Basics &
            $586,341$ &
            $11,726$ &
            $10.48\%$ \\
            
            \color{black!65} Locate avatar (LoA) &
            \color{black!65} $65,149$ &
            \color{black!65} $1,336$ &
            \color{black!65} $1.19\%$ \\
            
            \color{black!65} Locate goal (LoG) &
            \color{black!65} $65,149$ &
            \color{black!65} $1,273$ &
            \color{black!65} $1.14\%$ \\
            
            \color{black!65} Apply action (Act) &
            \color{black!65} $195,447$ &
            \color{black!65} $3,930$ &
            \color{black!65} $3.51\%$ \\
            
            \color{black!65} Sense condition (Sense) &
            \color{black!65} $260,596$ &
            \color{black!65} $5,187$ &
            \color{black!65} $4.64\%$ \\
            \midrule
            
            Fine-grained: Tracing &
            $15,152$ &
            $15,152$ &
            $13.54\%$ \\
            
            \color{black!65} Sequence trace &
            \color{black!65} $7,576$ &
            \color{black!65} $7,576$ & 
            \color{black!65} $6.77\%$ \\
            
            \color{black!65} Code trace &
            \color{black!65} $7,576$ &
            \color{black!65} $7,576$ &
            \color{black!65} $6.77\%$ \\
            \midrule
            
            Fine-grained: Grid synthesis &
            $68,184$ &
            $68,184$ &
            $60.95\%$ \\
            
            \color{black!65} Place avatar &
            \color{black!65} $7,576$ &
            \color{black!65} $7,576$ &
            \color{black!65} $6.77\%$ \\
            
            \color{black!65} Place goal &
            \color{black!65} $7,576$ &
            \color{black!65} $7,576$ &
            \color{black!65} $6.77\%$ \\
            
            \color{black!65} Place avatar+goal &
            \color{black!65} $7,576$ &
            \color{black!65} $7,576$ & 
            \color{black!65} $6.77\%$ \\
            
            \color{black!65}Place walls &
            \color{black!65} $37,880$ &
            \color{black!65} $37,880$ &
            \color{black!65} $33.87\%$ \\
            
            \color{black!65}Design all &
            \color{black!65} $7,576$ &
            \color{black!65} $7,576$ &
            \color{black!65} $6.77\%$ \\
            \midrule
            \textbf{Total} & \textbf{586,341} & \textbf{111,861} & \textbf{100\%} \\
            \bottomrule   
        \end{tabular}
        }
    \caption{Distributions for synthetically generated data.}
    \label{fig.data.distribution.table}
    \end{subfigure}
    \ \ \ \ \ \ \ \ \ \
    \begin{subfigure}[b]{.39\textwidth}
	\centering
        \includegraphics[width=.98\textwidth, trim={0cm 0.1cm 0cm 0cm},clip]{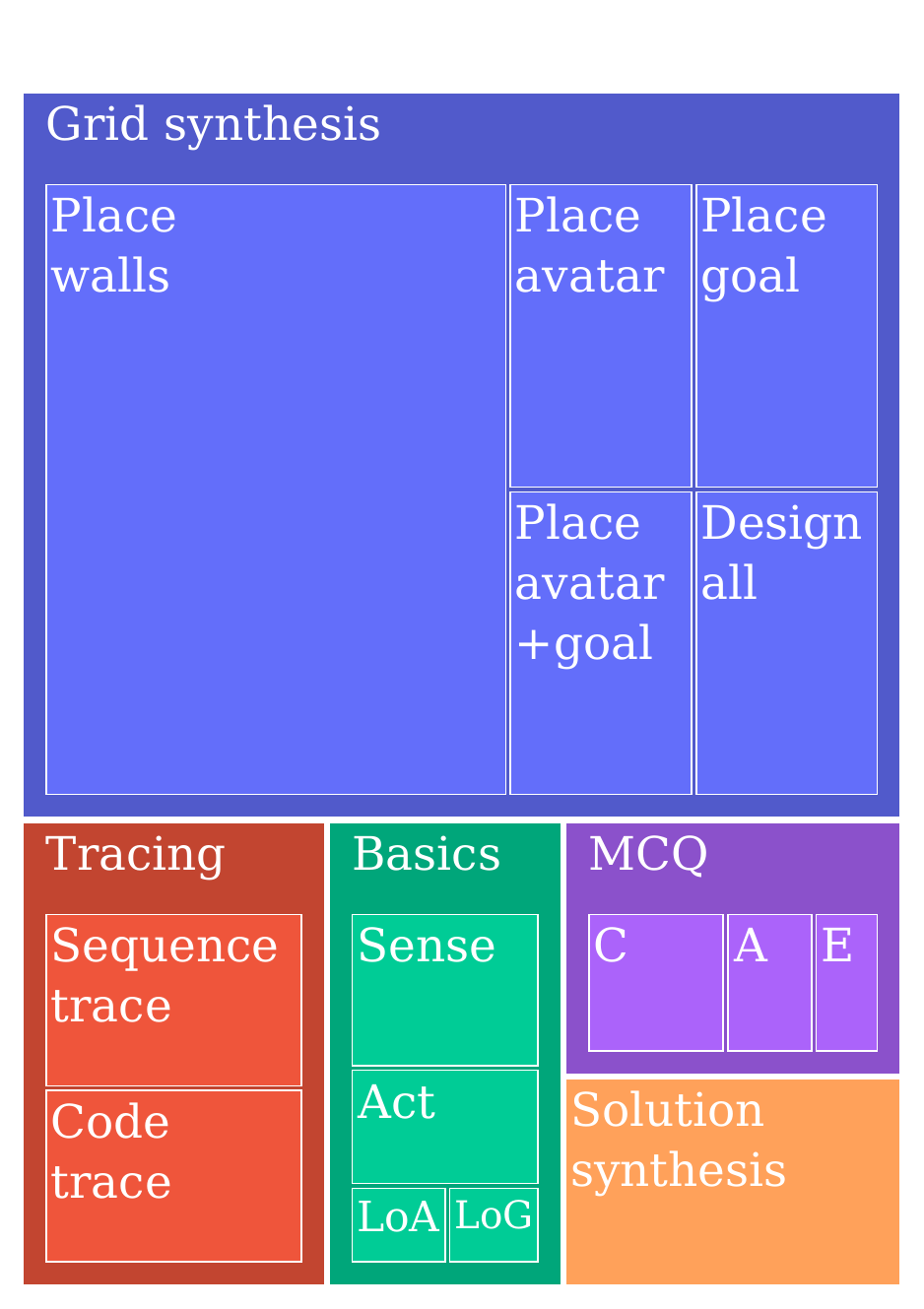}
        \vspace{-1.5mm}
        \caption{Treemap of selected data distribution.}
        \label{fig.data.distribution.treemap}
    \end{subfigure}
    \vspace{-1.5mm}
    \caption{Our synthetically generated training dataset. Subsampling is done only in the case of basics.
    }
    \label{fig.data.distribution}
    \vspace{-3mm}
\end{figure*}

\subsection{Synthetic Data for Multi-choice Programming Questions}
\label{sec.dataset.mcq}

We now focus on generating MCQ tasks similar to those in \aceType{} and \ctType{} \cite{DBLP:conf/sigcse/GhoshMS24,gonzalez2015computational,DBLP:journals/chb/Roman-GonzalezP17}. We generate MCQs starting from the same \codeGridSetPairs{} used for generating solution synthesis tasks, using a template-based approach, with manually written text descriptions for each task type. 
Next, we present our task types covering all the higher cognitive levels in Bloom's taxonomy -- \Analyzing{}, \Evaluating{}, \Creating{} \cite{DBLP:conf/sigcse/GhoshMS24,bloom1956taxonomy,krathwohl2002revision}. We also augment MCQs with explanations similar to the ones we use for solution synthesis tasks.

\textbf{\Analyzing{}.}
\looseness-1First, we describe the process of generating \Analyzing{} cognitive level tasks. For this level, we generate three task types: tasks that require selecting a solution code for a given grid, tasks that require indicating which given grids are solved by a given code, and tasks that require reasoning about the trace of a given code on a given grid and selecting the cells visited by the avatar. To offer an overview of our method, we explain the generation process of one task type, namely reasoning about the trace. We start from a pair \codeGridPair{} and the text description specific to this type of task. Then, we generate the correct option by executing \code{} on \grid{} and selecting random cells visited by the avatar. We generate distractor options by randomly picking free cells that were not visited by the avatar. Note that this task is correct by construction, unlike some more complex task types below that need validation. Finally, we have the task containing text description, \code{}, \grid{}, the correct option, and three distractor options.

\textbf{\Evaluating{}.}
\looseness-1Second, we describe the process of generating \Evaluating{} cognitive level tasks. For this level, we generate four task types: tasks that require identifying bugs, tasks that require repairing bugs, tasks that require evaluating code equivalence with no given grid, and tasks that require evaluating code equivalence given a grid. We explain generation process for the task type that requires repairing bugs. We start from pair \codeGridPair{} and corresponding text description. We generate a mutation and apply it to code \code{} to obtain $\code_{\text{mut}}$ \cite{DBLP:conf/nips/AhmedCEFGRS20,DBLP:conf/aied/GhoshTDS22}. The correct option is obtained as the reverse mutation that would transform $\code_{\text{mut}}$ back to \code{}. Distractor options are obtained by generating three other mutations. We validate the task by applying reverse mutation on $\code_{\text{mut}}$ and checking whether resulting code solves \grid{}. We also apply distractor mutations on $\code_{\text{mut}}$ and make sure that resulting codes do not solve \grid{}. Finally, we have the task containing text description, $\code_{\text{mut}}$, \grid{}, the correct option, and three distractor options.

\textbf{\Creating{}.}
\looseness-1Third, we describe the process of generating \Creating{} cognitive level tasks. For this level, we generate six types of tasks that require reasoning about modifying an incomplete grid such that the given code solves the modified grid. We generate: tasks that require placing the avatar, tasks that require reasoning about the number of possible initial avatar locations, tasks that require placing the goal, tasks that require counting possible goal positions, tasks that require placing walls, and tasks that require counting the minimum number of walls needed. For example, a task similar to Figure~\ref{fig.illustration.ace} can be synthesized by starting from a pair \codeGridPair{} and the text description. We set the correct option by randomly picking the number of walls to remove from $\{1, 2, 3\}$, in this case $1$. We remove one wall from \grid{}, obtaining $\grid_{\text{mut}}$. We generate three distractor options by applying arithmetic operations to the correct option. To validate the task correctness, we check whether \code{} solves any grid obtained via adding all possible combinations of walls less than the correct option. For this specific example, as the correct option for the example is $1$, we just need to check if the grid will be solved with no added walls.
Finally, we have the task containing text description, \code{}, $\grid_{\text{mut}}$, the correct option, and three distractor options.

\vspace{-1mm}
\subsection{Synthetic Data at Fine-grained Skills}
\label{sec.dataset.aug}
\vspace{-1mm}

We now introduce new kinds of tasks, aimed at improving fine-grained skills fundamental to making the models better understand the domains of our computational thinking tests. The main intuition for using various fine-grained skills is due to inter-task transfer observed during instruction-tuning \cite{DBLP:conf/iclr/WeiBZGYLDDL22,DBLP:conf/acl/0001LX22,DBLP:conf/acl/MokDLTYY23}, which can enhance performance on solution synthesis tasks and MCQ tasks. We now give details about generating three kinds of tasks for improving fine-grained skills.

\begin{figure*}
    \begin{subfigure}{0.49\textwidth}
        \centering
        \setlength{\fboxsep}{0.05pt}\fbox{\includegraphics[width=0.92\textwidth]{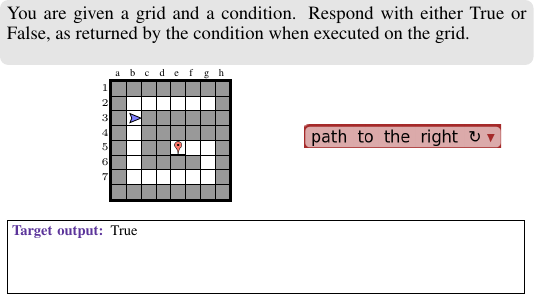}}
        \vspace{-1mm}
        \caption{Example for sense condition in basics.}
        \vspace{1mm}
        \label{fig.dataset.illustration.sensing}
        \end{subfigure}
    \begin{subfigure}{0.49\textwidth}
        \centering
        \setlength{\fboxsep}{0.05pt}\fbox{\includegraphics[width=0.92\textwidth]{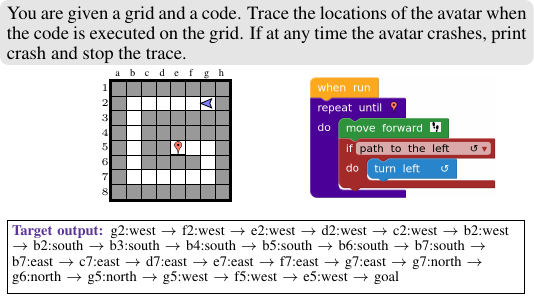}}
        \vspace{-1mm}
        \caption{Example for code trace in tracing.}
        \vspace{1mm}
        \label{fig.dataset.illustration.tracing}
    \end{subfigure}
    \\
    \begin{subfigure}{0.49\textwidth}
        \centering
        \setlength{\fboxsep}{0.05pt}\fbox{\includegraphics[width=0.92\textwidth]{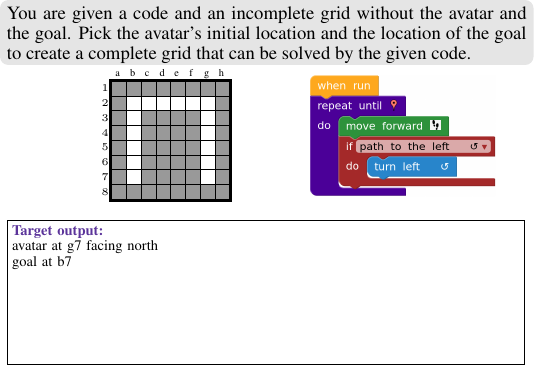}}
        \vspace{-1mm}
        \caption{Example for place avatar+goal in grid synthesis.}
        \vspace{1mm}
        \label{fig.dataset.illustration.gridsyn-avatar-goal}
    \end{subfigure}
    \begin{subfigure}{0.49\textwidth}
        \centering
        \setlength{\fboxsep}{0.05pt}\fbox{\includegraphics[width=0.92\textwidth]{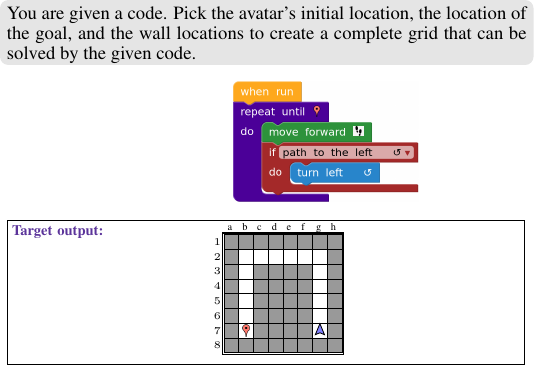}}
        \vspace{-1mm}
        \caption{Example for design all in grid synthesis.}
        \vspace{1mm}
        \label{fig.dataset.illustration.gridsyn-complete}
    \end{subfigure}
    \vspace{-2mm}    
    \caption{Illustrative examples for synthetically generated tasks with target outputs for the fine-grained skills. The tasks have been adapted for readability. For example, in \textbf{(d)}, we illustrate the target output visually, but the actual target output is in textual form.}
    \label{fig.dataset.illustration}
    \vspace{-4.5mm}    
\end{figure*}

\textbf{Basics.}
\looseness-1We describe the process of generating tasks aimed at familiarizing models with the fundamental aspects of visual programming. We generate four types of basic tasks: tasks that require locating the avatar in a given grid, tasks that require locating the goal in a given grid, tasks that require specifying the new location of the avatar after executing a given basic action on a given grid, and tasks which require specifying the outcome of applying a given condition to a given grid. For example, we generate the input for the task in Figure~\ref{fig.dataset.illustration.sensing} starting from a grid \grid{}, a randomly selected condition present in the DSL, and a fixed text description. The target output is obtained as the outcome of applying the condition on \grid{}, in this case True as the avatar has a free cell to its right. As the number of obtainable basic tasks is very large, we subsample to 2\% of the original size (see Figure~\ref{fig.data.distribution.table}). We have empirically chosen this percentage, analyzing the performance on a validation segment corresponding to basic tasks. 

\textbf{Tracing.}
Next, we describe the process of generating tasks aimed at enhancing the model's understanding of the interaction between the basic actions, conditions, and grids, crucial for answering the MCQs corresponding to the \Analyzing{} cognitive level in tests. We generate two types of tasks: tasks requiring to produce the trace obtained by applying a sequence of basic actions to a given grid, and tasks requiring to produce the trace of a given code on a given grid. For example, in Figure~\ref{fig.dataset.illustration.tracing}, we use a pair \codeGridPair{} and a fixed text description as input and treat the trace of \code{} on \grid{} as the target output. 

\textbf{Grid synthesis.}
\looseness-1Finally, we describe the generation of tasks aimed at boosting the model's understanding of the role of each grid element and how it can influence the execution of a code, crucial for answering the MCQs corresponding to the \Creating{} cognitive level in tests. 
We generate five types of tasks: tasks that require placing the avatar, tasks that require placing the goal, tasks that require placing both the avatar and the goal, tasks that require placing walls, and tasks that require designing a full grid. For example, we generate the task in Figure~\ref{fig.dataset.illustration.gridsyn-avatar-goal} starting from a pair \codeGridPair{} and a fixed text description, removing the avatar and the goal from \grid{}, and giving the incomplete grid and \code{} as input. The target output is the avatar and goal positions from \grid{}. Similarly, in Figure~\ref{fig.dataset.illustration.gridsyn-complete}, we keep only \code{} as input, and require as output a natural language description of \grid{}. We also include tracing information and sequences of basic actions as explanations during fine-tuning in a style similar to solution synthesis and MCQ tasks.

\section{Experiments} \label{sec.experiments}

\looseness-1In this section, we compare performance of open-access models, OpenAI's GPT family of models, and our fine-tuned \TechLlamaCT{}. We also include school students' performance based on studies in existing literature \cite{DBLP:conf/sigcse/GhoshMS24,DBLP:journals/chb/Roman-GonzalezP17}. We evaluate \TechLlamaCT{} variants to assess the impact of training on different data segments and the use of explanations. We also provide insights into models' reasoning process.

\subsection{Techniques Evaluated}

We start with \TechChance{} technique, a baseline that generates random tokens for \hocType{} tasks and selects most frequent answer from four options for \aceType{} and \ctType{} tasks. Note that because of non-uniform distribution of options, \TechChance{} yields better results for \aceType{} and \ctType{} than randomly choosing an option. Next, we present techniques based on generative models and performance of school students. Figure~\ref{fig.experiments.techniques} shows a summary of our model-based techniques.

\textbf{Open-access models.} We select smaller, instruction-tuned models from the Llama family, such as the 7B parameter version of CodeLlama~\cite{DBLP:journals/corr/abs-2308-12950} and the 8B parameter version of Llama3~\cite{Llama3}, alongside the 7B parameter version of Llava~\cite{DBLP:conf/nips/LiuLWL23a}. These are referred to as \TechLlamaSeven{}, \TechLlamaThreeEight{}, and \TechLlava{}, respectively. For \TechLlava{}, we incorporate both natural language and visual representations of grids to utilize its vision capabilities. All techniques are prompted to use chain-of-thought (CoT) \cite{DBLP:conf/nips/Wei0SBIXCLZ22}.

\begin{figure*}[t!]
\centering
    \scalebox{0.8}{
        \setlength\tabcolsep{4pt}
        \aboverulesep=0pt
        \belowrulesep=0pt
        \renewcommand{\arraystretch}{1.3}
        \begin{tabular}{llccll}  
            \toprule
            \multirow{1}{*}{\textbf{Technique}} &
            \multirow{1}{*}{\textbf{Base model}} &
            \multicolumn{2}{c}{\textbf{Modality}} &
            \multirow{1}{*}{\textbf{Fine-tuning data}} &
            \multirow{1}{*}{\textbf{Explanation}} \\
            \cmidrule(lr){3-4}

            & & Visual & Text & & \\
            
            \midrule
            $\TechLlamaSeven$ &
            CodeLlama-7B \cite{DBLP:journals/corr/abs-2308-12950} &
            \notick{} &
            \yestick{} &
            n/a &
            n/a \\

            $\TechLlava$ &
            LLaVA-v1.5-7B \cite{DBLP:conf/nips/LiuLWL23a} &
            \yestick{} &
            \yestick{} &
            n/a &
            n/a \\

            $\TechLlamaThreeEight$ &
            Llama3-8B \cite{Llama3} &
            \notick{} &
            \yestick{} &
            n/a &
            n/a \\

            \midrule
            \multirow{1}{*}{$\TechChatGPT$, $\TechGPTFour$, $\TechGPTFourO$} &
            \multirow{1}{*}{GPT-3.5 \cite{ChatGPT}, 4 \cite{GPT4}, 4o \cite{GPT4o}} &
            \multirow{1}{*}{\notick{}} &
            \multirow{1}{*}{\yestick{}} &
            \multirow{1}{*}{n/a} &
            \multirow{1}{*}{n/a} \\

            \multirow{1}{*}{$\TechGPTFourV$, $\TechGPTFourOVision$} &
            \multirow{1}{*}{GPT-4V \cite{GPT4V}, 4o \cite{GPT4o}} &
            \multirow{1}{*}{\yestick{}} &
            \multirow{1}{*}{\notick{}} &
            \multirow{1}{*}{n/a} &
            \multirow{1}{*}{n/a} \\

            \multirow{1}{*}{$\TechGPTFourVCombined$, $\TechGPTFourOCombined$} &
            \multirow{1}{*}{GPT-4V \cite{GPT4V}, 4o \cite{GPT4o}} &
            \multirow{1}{*}{\yestick{}} &
            \multirow{1}{*}{\yestick{}} &
            \multirow{1}{*}{n/a} &
            \multirow{1}{*}{n/a} \\

            \midrule
            $\TechHocFinetuneNoExpThree$ &
            Llama3-8B \cite{Llama3} &
            \notick{} &
            \yestick{} &
            Solution syn &
            None \\

            $\TechPlainFinetuneThree$ &
            Llama3-8B \cite{Llama3} &
            \notick{} &
            \yestick{} &
            Solution syn+MCQ &
            None \\

            $\TechHocFinetuneExpThree$ &
            Llama3-8B \cite{Llama3} &
            \notick{} &
            \yestick{} &
            Solution syn &
            Train \\

            $\TechTopFinetuneThree$ &
            Llama3-8B \cite{Llama3} &
            \notick{} &
            \yestick{} &
            Solution syn+MCQ &
            Train \\

            $\TechHierFinetuneThree$ &
            Llama3-8B \cite{Llama3} &
            \notick{} &
            \yestick{} &
            Full data &
            Train \\

            $\TechHierEmuFinetuneEmuInfThree$ &
            Llama3-8B \cite{Llama3} &
            \notick{} &
            \yestick{} &
            Full data &
            Train+infer \\

            \bottomrule   
        \end{tabular}
        }
    \caption{Table summarizing techniques based on generative models, showing the base model and whether the input grid is represented visually or in text (modality). For fine-tuned models (e.g., \TechLlamaCT{}), the table specifies the data segment used for training and whether models were trained with no explanations, to generate explanations, or to receive explanations during inference.}
    \label{fig.experiments.techniques}
    \vspace{-4mm}
\end{figure*}

\looseness-1\textbf{GPT family.} This group includes techniques based on GPT-3.5 \cite{ChatGPT} and GPT-4 \cite{GPT4o,GPT4}. We start with \TechChatGPT{} technique which processes tasks, including grids, only in natural language, as it has no vision capabilities. Similarly, \TechGPTFour{} is solely based on natural language. Next, for the \TechGPTFourV{}, we input the grids solely as visual representation, while the rest of the task is represented through natural language. \TechGPTFourVCombined{} technique combines textual and visual representations for grids, with the rest of the task in natural language. We also include similar techniques based on the newer GPT-4o~\cite{GPT4o}, namely \TechGPTFourO{}, \TechGPTFourOVision{}, and \TechGPTFourOCombined{}. All techniques are prompted to use CoT.\footnote{Few-shot prompting did not improve results. All results are based on zero-shot CoT prompting.}

\textbf{Fine-tuned models.} We fine-tune the instruction-tuned 8B parameter version of the Llama3 model using LoRA~\cite{DBLP:conf/iclr/HuSWALWWC22} and obtain the \TechLlamaCT{} family. \TechHocFinetuneNoExpThree{} and \TechHocFinetuneExpThree{} are fine-tuned using only the generated solution synthesis tasks, \TechPlainFinetuneThree{} and \TechTopFinetuneThree{} are trained using both generated solution synthesis tasks and generated MCQ tasks, and \TechHierFinetuneThree{} is trained on the full synthetic dataset. \TechHocFinetuneExpThree{}, \TechTopFinetuneThree{}, and \TechHierFinetuneThree{} are trained on target outputs enriched with explanations. Additionally, \TechHierEmuFinetuneEmuInfThree{} simulates an ideal scenario where the correct reasoning process is known at inference time.

\textbf{Human students.} We benchmark these models against the performance of students observed in literature, reporting results for one group of students for \hocType{} and \aceType{} \cite{DBLP:conf/sigcse/GhoshMS24}, and for a different group of students for \ctType{} \cite{DBLP:journals/chb/Roman-GonzalezP17}. \TechHumanAll{} comprises the average performance of students across grades 3-7 for \hocType{} and \aceType{}, and the average performance of students across grades 5-10 for \ctType{}. \TechHuman{} represents the top $25\%$ of grade 7 students for \hocType{} and \aceType{}, and the top $25\%$ of grade 7-8 students for \ctType{}, showing the performance of the best students.

\subsection{Performance on Computational Thinking Tests}

\looseness-1We evaluate techniques on \hocType{}, \aceType{}, and \ctType{}, introduced in Section~\ref{sec.background.tests}. Figure~\ref{fig.experiments.results_table} shows results, with accuracy computed as percentage of correctly answered tasks in one trial out of total tasks per test. We set temperature to 0 and assess over three seeds, reporting average results as mean (stderr).

\textbf{Combining language and vision enhances performance.}
Providing input in both text and visual modality leads to better results for \TechGPTFourOCombined{} when compared with \TechGPTFourOVision{} and \TechGPTFourO{}. Similar results hold for \TechGPTFourVCombined{} when compared with \TechGPTFourV{} and \TechGPTFour{}.

\begin{figure*}[t!]
\centering
    \scalebox{0.8}{
        \setlength\tabcolsep{18pt}
        \renewcommand{\arraystretch}{1}
        \begin{tabular}{lrrrr}  
            \toprule
             \multirow[c]{1}{*}[0mm]{\textbf{Technique}} & 
             \multicolumn{1}{c}{\hocTypeBold{}} &
             \multicolumn{1}{c}{\aceTypeBold{}} &
             \multicolumn{1}{c}{\ctTypeBold{}} &
             \multicolumn{1}{c}{\textbf{Overall}} \\

            \midrule
            $\TechChance$ &
            $0.0$ $\phantom{(1.1)}$&
            $33.0$ $\phantom{(1.1)}$&
            $33.0$ $\phantom{(1.1)}$&
            $22.0$ $\phantom{(1.1)}$\\
            
            \midrule
            $\TechLlamaSeven$ &
            $0.0$ $(0.0)$&
            $14.3$ $(0.0)$&
            $29.2$ $(0.0)$&
            $14.3$ $(0.0)$\\

            $\TechLlava$ &
            $0.0$ $(0.0)$&
            $28.6$ $(0.0)$&
            $20.8$ $(0.0)$&
            $16.7$ $(0.0)$\\

            \rowcolor{red!15}
            $\TechLlamaThreeEight$ &
            $0.0$ $(0.0)$&
            $34.9$ $(2.0)$&
            $34.7$ $(5.0)$&
            $22.9$ $(1.3)$\\

            \midrule
            $\TechChatGPT$ &
            $25.0$ $(0.0)$&
            $31.7$ $(4.0)$&
            $36.1$ $(5.0)$&
            $31.1$ $(0.5)$\\

            $\TechGPTFourV$ &
            $18.3$ $(2.0)$&
            $31.7$ $(8.0)$&
            $44.4$ $(3.0)$&
            $31.6$ $(1.3)$\\

            $\TechGPTFour$ &
            $21.7$ $(2.0)$&
            $52.4$ $(6.0)$&
            $56.9$ $(5.0)$&
            $43.7$ $(2.9)$\\

            $\TechGPTFourVCombined$ &
            $28.3$ $(2.0)$&
            $57.1$ $(3.0)$&
            $58.3$ $(5.0)$&
            $48.0$ $(0.4)$\\

            $\TechGPTFourOVision$ &
            $20.0$ $(0.0)$&
            $38.1$ $(3.0)$&
            $52.8$ $(3.0)$&
            $36.9$ $(1.6)$\\

            $\TechGPTFourO$ &
            $30.0$ $(0.0)$&
            $61.9$ $(3.0)$&
            $59.7$ $(3.0)$&
            $50.7$ $(1.7)$\\

            \rowcolor{orange!15}
            $\TechGPTFourOCombined$ &
            $30.0$ $(0.0)$&
            $61.9$ $(0.0)$&
            $66.7$ $(0.0)$&
            $53.0$ $(0.0)$\\
            
            \midrule
            $\TechHocFinetuneNoExpThree$&
            $10.0$ $(0.0)$&
            $30.5$ $(2.0)$&
            $25.0$ $(0.0)$&
            $21.9$ $(0.7)$\\
            
            $\TechPlainFinetuneThree$&
            $11.7$ $(4.0)$&
            $44.4$ $(5.0)$&
            $33.3$ $(5.0)$&
            $29.8$ $(1.2)$\\

            $\TechHocFinetuneExpThree$&
            $55.0$ $(4.0)$&
            $27.6$ $(3.0)$&
            $23.1$ $(2.0)$&
            $35.3$ $(0.9)$\\
            
            $\TechTopFinetuneThree$&
            $40.0$ $(9.0)$&
            $43.5$ $(3.0)$&
            $36.1$ $(2.0)$&
            $40.0$ $(3.6)$\\

            \rowcolor{green!60!black!15}
            $\TechHierFinetuneThree$&
            $50.0$ $(4.0)$&
            $57.8$ $(1.0)$&
            $51.4$ $(3.0)$&
            $53.0$ $(1.7)$\\
            
            $\TechHierEmuFinetuneEmuInfThree$& 
            $76.7$ $(7.0)$&
            $74.6$ $(4.0)$&
            $65.3$ $(0.0)$&
            $72.2$ $(1.3)$\\

            \midrule
            \rowcolor{cyan!15}
            $\TechHumanAll$ &
            $74.1$ $\phantom{(1.1)}$&
            $50.9$ $\phantom{(1.1)}$&
            $58.5$ $\phantom{(1.1)}$ &
            $61.2$ $\phantom{(1.1)}$\\

            \rowcolor{tabblue!20}
            $\TechHuman$ &
            $99.8$ $\phantom{(1.1)}$&
            $84.0$ $\phantom{(1.1)}$&
            $71.4$ $\phantom{(1.1)}$ &
            $85.1$ $\phantom{(1.1)}$\\

            \bottomrule   
        \end{tabular}
        }
    \caption{Results on \hocType{}, \aceType{}, \ctType{}, and overall performance.}
    \label{fig.experiments.results_table}
\end{figure*}
\begin{figure*}[t!]
\centering
    \scalebox{0.8}{
        \setlength\tabcolsep{7pt}

        \renewcommand{\arraystretch}{1.2}
        \begin{tabular}{lrrrrrr}  
            \toprule
             \multirow[c]{1}{*}[0mm]{\textbf{Technique}} & 
             \multicolumn{1}{c}{\hocTypeBold{}} &
             \multicolumn{1}{c}{\hocTypeBold{}} &
             \multicolumn{1}{c}{\aceTypeBold{}} &
             \multicolumn{1}{c}{\aceTypeBold{}} &
             \multicolumn{1}{c}{\ctTypeBold{}} &
             \multicolumn{1}{c}{\ctTypeBold{}} \\

             & & \textbf{reasoning} & & \textbf{reasoning} & & \textbf{reasoning} \\
             \midrule

            $\TechLlamaThreeEight$ &
            $0.0$ $(0.0)$&
            $4.2$ $(1.0)$&
            $34.9$ $(2.0)$&
            $4.8$ $(0.0)$&
            $34.7$ $(5.0)$&
            $0.0$ $(0.0)$\\

            $\TechGPTFourOCombined$ &
            $30.0$ $(0.0)$&
            $35.0$ $(0.0)$&
            $61.9$ $(0.0)$&
            $28.6$ $(0.0)$&
            $66.7$ $(0.0)$&
            $39.6$ $(0.0)$\\

            $\TechHierFinetuneThree$&
            $50.0$ $(4.0)$&
            $73.3$ $(8.0)$&
            $57.8$ $(1.0)$&
            $32.5$ $(3.0)$&
            $51.4$ $(3.0)$&
            $23.6$ $(3.0)$\\
            
            \bottomrule   
        \end{tabular}
        }

    \caption{Comparison of accuracy in correctly answered tasks and reasoning correctness across domains for representative models in \hocType{}, \aceType{}, and \ctType{} tasks, reported as mean (stderr). Reasoning correctness results are based on manual annotations done by two independent annotators.}
    \label{fig.reasoning.annotations}
    \vspace{-3mm}
\end{figure*}

\textbf{Symbolic information-based explanations improve outcomes.}
\looseness-1Template-based explanations derived from execution information used while training enhance reasoning at inference and boost model performance. Specifically, \TechHocFinetuneExpThree{} and \TechTopFinetuneThree{}, which are trained with explanations, outperform their counterparts \TechHocFinetuneNoExpThree{} and \TechPlainFinetuneThree{} trained only for generating an answer with no explanation.

\textbf{Fine-grained skills make \TechLlamaCT{} comparable to GPT-4.}
We notice an increase of at least $10\%$ in performance on \hocType{}, \aceType{}, and \ctType{} for the model fine-tuned with fine-grained skills data. Fine-tuning with explanations and across the full dataset allows \TechHierFinetuneThree{} to achieve overall results comparable to those of \TechGPTFourOCombined{}. This shows that a better understanding of the visual domain is key to better performance on all three tests.

\textbf{Reasoning for MCQ tasks is harder than for solution synthesis tasks.} We analyze the reasoning capabilities of three selected models, $\TechLlamaThreeEight$, $\TechGPTFourOCombined$, and $\TechHierFinetuneThree$, through manual annotations of their reasoning process. A model's reasoning is considered correct if it respects grid constraints, correctly maps codes to traces and sequences, and avoids introducing unnecessary details. We follow a strict binary metric, where the reasoning process is marked as correct (i.e., $1$) if the entire reasoning process is correct and incorrect (i.e., $0$) otherwise. The reasoning processes were reviewed by two independent annotators\footnote{The annotators obtained a Cohen's kappa score of $0.84$, indicating high agreement \cite{Cohen1960ACO}.}. Figure~\ref{fig.reasoning.annotations} compares accuracy in correctly answered tasks with reasoning correctness averaged across annotators and aggregated over three seeds, for \hocType{}, \aceType{}, and \ctType{} tasks. $\TechLlamaThreeEight$ often tries guessing answers without providing any reasoning, while $\TechGPTFourOCombined$ struggles with grid layouts, sometimes missing walls. $\TechHierFinetuneThree$ traces tasks well in \hocType{} but faces challenges with converting sequences to minimal codes and tracing in \aceType{} and \ctType{}.

\textbf{Symbolic information at inference leads to human-level performance.}
Including explanations with a correct reasoning process in the input prompts increases performance, bringing it closer to that of school students. However, \TechHierEmuFinetuneEmuInfThree{} simulates an ideal scenario, as correct explanations are usually not available at inference as input.

\textbf{Human students are better at solution synthesis.}
\looseness-1Figure~\ref{fig.experiments.comp.radar_tests} showcases that state-of-the-art and fine-tuned models have slightly better performance than the average grade 3-7 student across three analyzed levels of Bloom's taxonomy, and that state-of-the-art models struggle with solution synthesis. Figure~\ref{fig.experiments.comp.radar_concepts} shows a deeper analysis of performance on \hocType{}, breaking down the performance per concept. It shows that by fine-tuning, a model's understanding of programming concepts grows similarly to that of an average student. Finally, Figure~\ref{fig.experiments.comp.scatter} compares models' performance on \hocType{} and \aceType{} tests with that of students from various grades. It reveals that models have not yet reached the problem-solving capabilities of grade 3 students on \hocType{} tasks. Besides spatial reasoning, adhering to constraints such as the required size and constructs is another reason for this weak performance. Interestingly, models can match the performance of grade 7 students on \aceType{} tests, where answer options are available.

\begin{figure}[t!]
    \centering
        \begin{subfigure}{\textwidth}
            \centering
                \includegraphics[trim={0cm 0cm 0cm 0cm},clip,width=\textwidth]{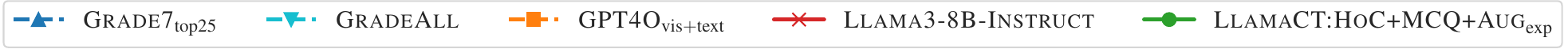}
            \vspace{-1mm}
        \end{subfigure}
    \\ 
    \begin{minipage}{0.5\textwidth}
    \centering
        \begin{subfigure}{0.7\textwidth}
            \centering
                \includegraphics[trim={0cm 0cm 0cm 0cm},clip,width=\textwidth]{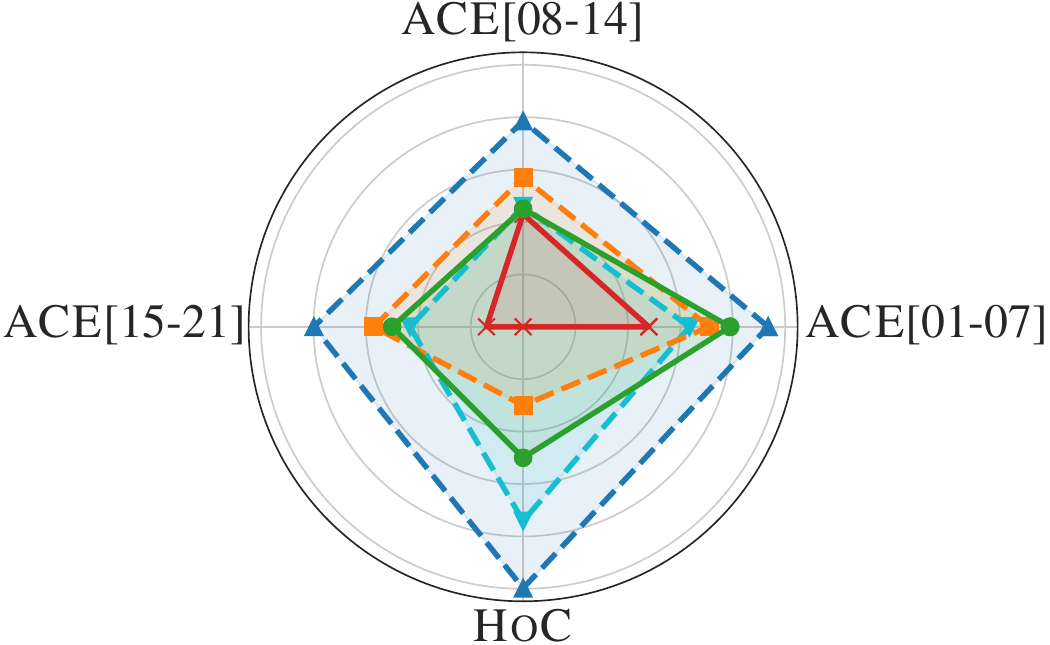}
                \vspace{-5mm}
            \caption{\hocType{} and \aceType{} levels.}
            \label{fig.experiments.comp.radar_tests}
            \vspace{3mm}
        \end{subfigure}
        \\
        \begin{subfigure}{0.7\textwidth}
                \hspace*{0.35cm}
                \includegraphics[trim={0cm 0cm 0cm 0cm},clip,width=\textwidth]{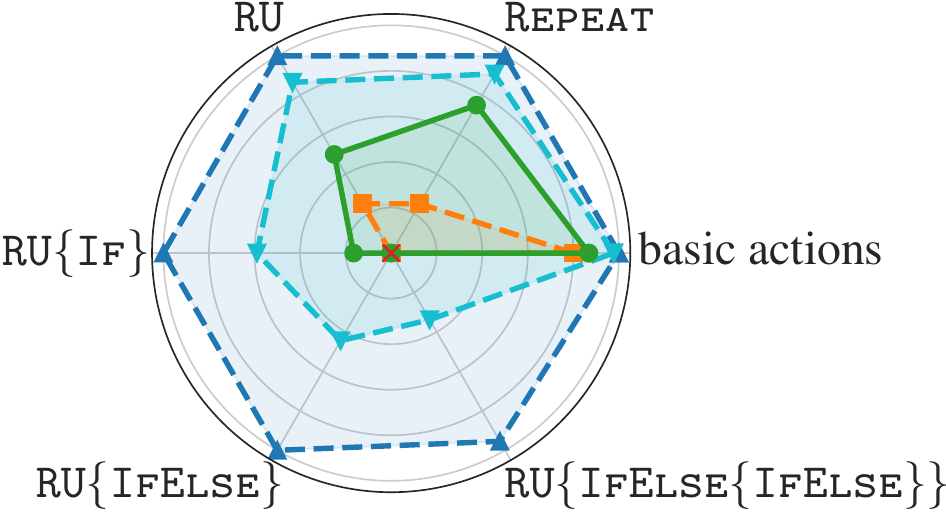}
            \caption{\hocType{} concepts.}
            \label{fig.experiments.comp.radar_concepts}
        \end{subfigure}
    \end{minipage}
    \begin{minipage}{0.47\textwidth}
    \centering
        \begin{subfigure}{0.99\textwidth}
            \centering
                \includegraphics[trim={0.3cm 0.3cm 0.3cm 0.3cm},clip,width=\textwidth]{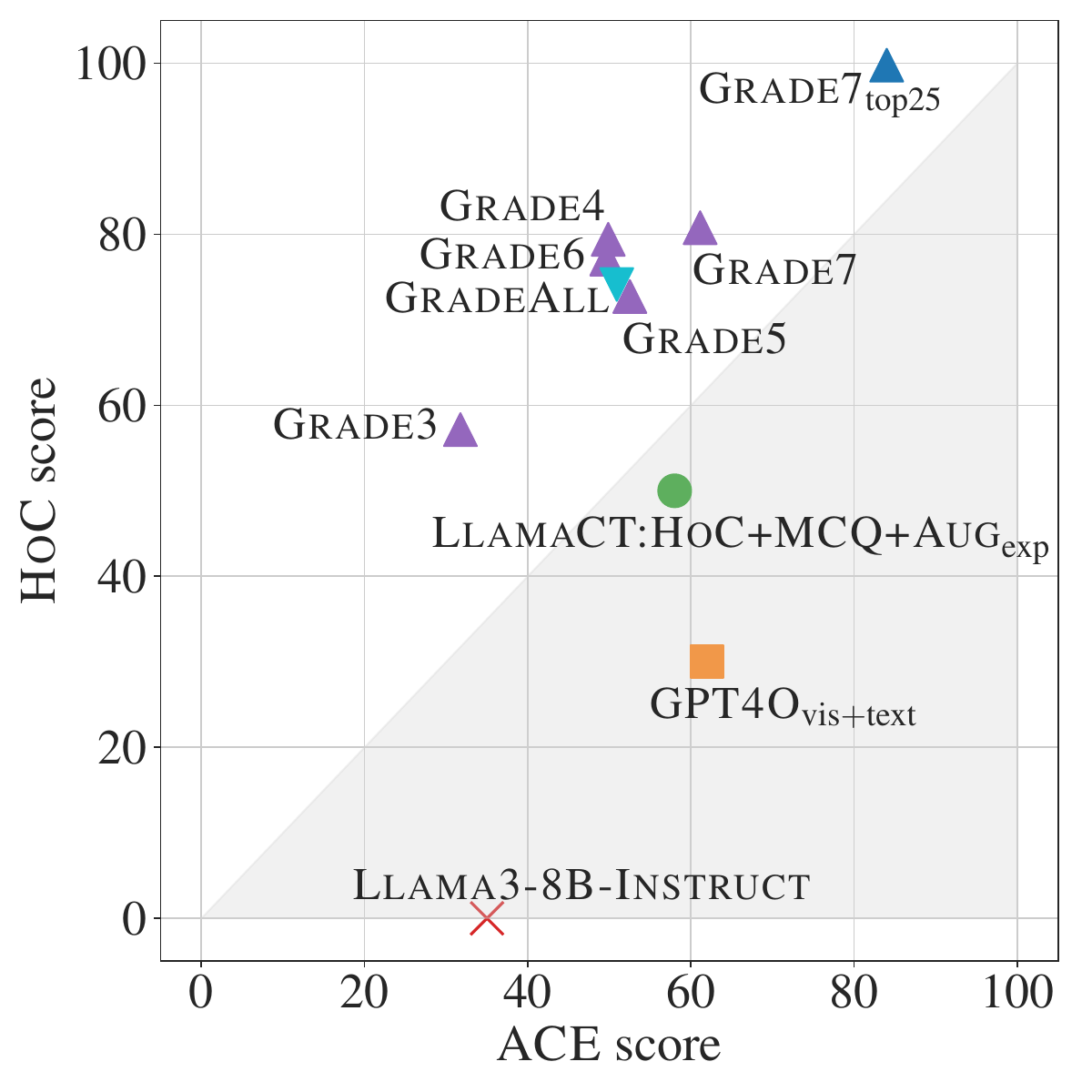}
            \caption{\hocType{} and \aceType{}.}
            \label{fig.experiments.comp.scatter}
        \end{subfigure}
    \end{minipage}
    \caption{\looseness-1Comparison between the performance of the best techniques and school students (grade 3-7) on a scale of $0$ to $100$.
    For better visualization and comparison, we present results only for \hocType{} and \aceType{}.
    \textbf{(a)} shows that state-of-the-art and fine-tuned models have a similar performance to an average grade 3-7 student on the \aceType{}, but lag behind for \hocType{}. \textbf{(b)} shows that fine-tuning can help models' problem-solving skills get closer to an average grade 3-7 student for simpler concepts. (\textcode{RU} stands for \DSLRepeatUntil{}). \textbf{(c)} shows how every grade is dominating models for \hocType{}. It also shows that state-of-the-art and fine-tuned models are close to the average grade 7 students' performance on \aceType{}. However, the performance of the best 25\% grade 7 students is still far from reach for generative models.}
    \label{fig.experiments.comp}
    \vspace{-3.5mm}
\end{figure}    
\vspace{-1mm}
\section{Concluding Discussion} \label{sec.conclusion}
\vspace{-2mm}

\looseness-1In this paper, we introduced a new benchmark for assessing generative models on computational thinking tests grounded in elementary visual programming. We made a detailed analysis of the performance of open-access models such as Llama3 and the GPT family of models, comparing it to that of school students. To boost performance of Llama3-8B, we fine-tuned it using our novel synthetic generation methodology based on symbolic information. The best fine-tuned model has a performance similar to state-of-the-art models, even though it is much smaller and does not use vision capabilities.

While our analysis gives a deep insight into the computational thinking and problem-solving capabilities of generative models, there are some limitations of our current work and directions to tackle them in future work. First, we assess multi-modal models on our benchmark but do not fine-tune them to improve performance. An interesting direction for future work is fine-tuning multi-modal models for solving computational thinking and problem-solving tasks. Second, one of our techniques naively uses correct explanations provided at inference time to help it reach an answer. An interesting direction for future work is developing techniques where generative models interact with symbolic tools to obtain this kind of information at inference time, possibly via multiple rounds of interaction. 
\subsubsection*{Acknowledgements}
Funded/Co-funded by the European Union (ERC, TOPS, 101039090). Views and opinions expressed are however those of the author(s) only and do not necessarily reflect those of the European Union or the European Research Council. Neither the European Union nor the granting authority can be held responsible for them.
    \bibliographystyle{unsrt}
    \bibliography{main}
    \clearpage
\appendix
{
    \allowdisplaybreaks
\section{Table of Contents}\label{app-sec.toc}
In this section, we provide a brief description of the content provided in the appendices of the paper.
\begin{itemize}[leftmargin=*]
    \item Appendix~\ref{app-sec.discussion} provides a discussion of the broader impact of our work, a responsibility statement, compute resources used, and training details.
    \item Appendix~\ref{app-sec.methodology} has a more detailed description of the generation methodology in \iftoggle{MainSuppContent}{Section~\ref{sec.dataset}}{Section~4}.
    \item Appendix~\ref{app-sec.data} gives more details about the origin of the curated data and more examples for each test.
    \item Appendix~\ref{app-sec.synth-data} details the synthetic dataset evaluation segment and shows results of selected techniques on this segment.
    \item Appendix~\ref{app-sec.reasoning} offers an insight into the reasoning process done by models.
    \item Appendix~\ref{app-sec.prompts} shows the templates of the prompts used for interacting with and training the models.
\end{itemize}
\section{Discussion}\label{app-sec.discussion}
\textbf{Broader impact.} 
This paper introduces a new benchmark for assessing the generative models' performance on computational thinking and problem-solving tasks. It also includes a dataset for potentially improving newer models (e.g., via fine-tuning). We believe our proposed benchmark has the potential to bring more attention to the problems state-of-the-art generative models encounter when it comes to computational thinking and problem-solving, thus leading to an improvement in their reasoning capabilities.

\textbf{Responsibility statement.} The authors declare that they bear full responsibility for any violations of rights, including but not limited to copyright infringement, plagiarism, or any other legal or ethical breaches, that may arise from the content and data provided in this work.

\textbf{Compute resources.} All the experiments were conducted on a cluster of machines equipped with Intel Xeon Platinum 8360Y CPUs running at a frequency of 2.40GHz and 8x NVIDIA A100 80GB PCIe GPUs.

\textbf{Training details.} We fine-tuned Llama3 using the llama-recipes repository\footnote{Found at \url{https://github.com/meta-llama/llama-recipes}}, using LoRA and FSDP with the default settings and hyperparameters, unless stated otherwise. We set LoRA $\text{r}=16$ and $\text{alpha}=32$. We pass as target modules for LoRA the following: ``q\_proj'', `v\_proj'', ``k\_proj'', ``o\_proj'', ``gate\_proj'', ``up\_proj'', ``down\_proj'', and ``lm\_head''.
We chose these LoRA parameters to strike a balance between accuracy, inference speed, and the size of the resulting LoRA adapter.
The data is naturally split into evaluation (i.e., the real-world benchmark) and training (i.e., the synthetic dataset). We further randomly split the synthetic dataset into train (90\%) and validation segments (10\%). We use the validation segment to check whether the loss is decreasing. As we have noticed that the decrease in loss is minimal after the second epoch, we always choose to train for 2 epochs. Training the 8B parameter version of Llama3 on the full synthetic dataset, with the resources described above takes approximately 10 hours. Training it on solution synthesis only takes approximately 1 hour, while training it on solution synthesis and multiple choice questions takes roughly 2 hours. Doing it for all the versions of \TechLlamaCT{} for multiple seeds leads to approximately 78-80 hours of compute. 
\clearpage
\section{Further Details about Data Generation}\label{app-sec.methodology}

In this section we give more insight into the data generation methodology described in \iftoggle{MainSuppContent}{Section~\ref{sec.dataset}}{Section~4}. We first give details about the MCQ task generation process, then we continue with the generation process for the fine-grained skills tasks.

\subsection{MCQ Tasks Generation}

We will now describe our generation methodology exemplified with two types of tasks.

\begin{figure*}[!h]
    \begin{subfigure}{0.99\textwidth}
        \centering
        \includegraphics[width=0.98\textwidth]{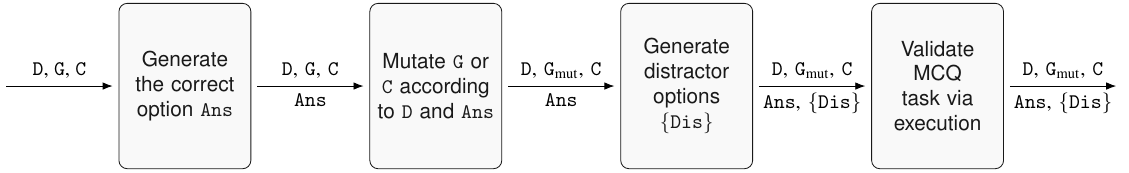}
        \vspace{-1mm}
        \caption{Illustration of the MCQ task generation methodology for minimum wall counting.}
        \vspace{2mm}
        \label{app-fig.mcq-pipeline-walls.pipeline}
    \end{subfigure}
    \\
    \begin{subfigure}{0.49\textwidth}
        \centering
        \includegraphics[width=0.98\textwidth]{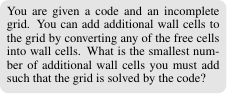}
        \vspace{-1mm}
        \caption{Manually written input text description \descr{}}
        \vspace{3mm}
        \label{app-fig.mcq-pipeline-walls.descr}
    \end{subfigure}
    \ \
    \begin{subfigure}{0.49\textwidth}
        \centering
        \includegraphics[width=0.98\textwidth]{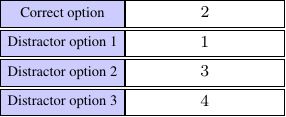}
        \vspace{-1mm}
        \caption{Answer options $\answer + \distractors$}
        \vspace{3mm}
        \label{app-fig.mcq-pipeline-walls.options}
    \end{subfigure}
    \\
    \begin{subfigure}{0.245\textwidth}
        \centering
        \includegraphics[width=0.98\textwidth]{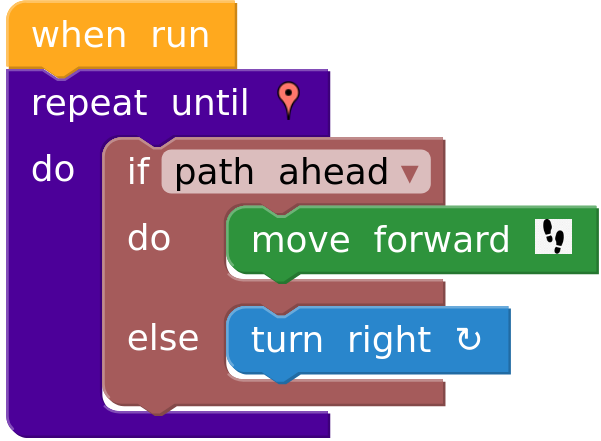}
        \vspace{1.5mm}
        \caption{Input code \code{}}
        \vspace{1mm}
        \label{app-fig.mcq-pipeline-walls.code}
    \end{subfigure}
    \begin{subfigure}{0.245\textwidth}
        \centering
        \includegraphics[width=0.98\textwidth]{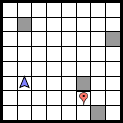}
        \vspace{-4mm}
        \caption{Input grid \grid{}}
        \vspace{1mm}
        \label{app-fig.mcq-pipeline-walls.grid}
    \end{subfigure}
    \begin{subfigure}{0.245\textwidth}
        \centering
        \includegraphics[width=0.98\textwidth]{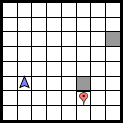}
        \vspace{-4mm}
        \caption{Mutated grid $\grid_{\text{mut}}$}
        \vspace{1mm}
        \label{app-fig.mcq-pipeline-walls.grid-mut}
    \end{subfigure}
    \begin{subfigure}{0.245\textwidth}
        \centering
        \includegraphics[width=0.98\textwidth]{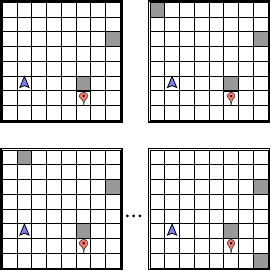}
        \vspace{-4mm}
        \caption{Combinations of walls}
        \vspace{1mm}
        \label{app-fig.mcq-pipeline-walls.valid}
    \end{subfigure}
    \vspace{-2mm}    
    \caption{Illustration of our MCQ task generation methodology instantiated on the minimum wall counting task in \Creating{} cognitive level. \textbf{(a)} gives an overview of the pipeline. \textbf{(b,d,e)} show the input components for the pipeline: text description \descr{}, grid \grid{}, and its solution code \code{}. \textbf{(c)} shows the generated correct option \answer{} along with the generated distractors \distractors{}, put together to obtain all four options. \textbf{(f)} shows the mutated grid $\grid_{\text{mut}}$ obtained according to the selected correct option \answer{}. \textbf{(g)} shows the grids obtained by combinatorially adding numbers of walls less than the correct option \answer{}. The MCQ task is valid if no combinatorially obtained grid can be solved by \code{}.}
    \label{app-fig.mcq-pipeline-walls}  
\end{figure*}


Figure~\ref{app-fig.mcq-pipeline-walls} shows an overview of our generation process for MCQ tasks. To give a better understanding, we will first exemplify the generation process for a task involving counting the minimum number of walls to add to the given grid so that the given code solves the modified grid. We start from a manually written text description for a task \descr{}, a grid \grid{}, and a code \code{}. \descr{} decides the type of task that will be generated and what operations need to be done at each step. The first step is to generate the correct answer option. This is selected from a pool of options which are either fixed or obtained via executing \code{} on \grid{}. In our example, we pick a number from $\{1, 2, 3\}$. Let us say that we have picked $\answer = 2$. We then proceed to the next step, namely mutating \code{} or \grid{} according to the task type and the picked correct option \answer{}. In our example, we will only mutate \grid{} by removing two randomly picked walls and obtain $\grid_{\text{mut}}$. Going on to the next step, we generate distractor options based on the information we have until now. For this example, we apply arithmetic operations to \answer{}, thus obtaining the set of distractors \distractors{}. We now have all the components of an MCQ task: the text description \descr{}, the mutated grid $\grid_{\text{mut}}$, code \code{}, the correct option \answer{}, and the three distractors \distractors{}. In the final step, we validate the task via execution. In our example, we check whether \code{} solves any grid obtained via adding all possible combinations of walls less than the correct option. We start by adding no walls and check whether \code{} solves the grid. Then we add one wall at a time to each free cell and check if \code{} solves the obtained grid. We discard the task in case any of the grids obtained via combinatorially adding walls is solved by \code{}. Otherwise, the task passes validation and we keep it.

\begin{figure*}[!h]
    \begin{subfigure}{0.99\textwidth}
        \centering
        \includegraphics[width=0.98\textwidth]{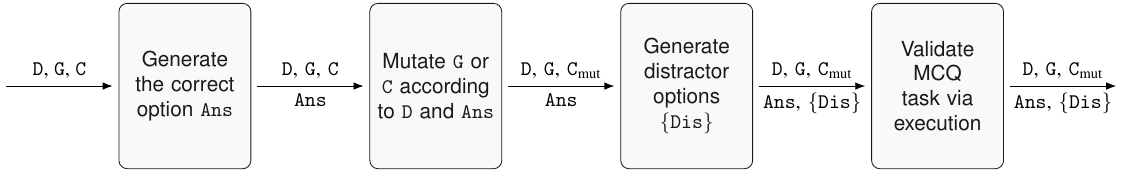}
        \vspace{-1mm}
        \caption{Illustration of the MCQ task generation methodology for bug repair.}
        \vspace{2mm}
        \label{app-fig.mcq-pipeline-bug.pipeline}
    \end{subfigure}
    \\
    \ \ \ \ \
    \begin{subfigure}{0.3\textwidth}
        \centering
        \includegraphics[width=0.98\textwidth]{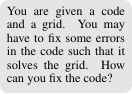}
        \vspace{-1mm}
        \caption{Input text description \descr{}}
        \vspace{3mm}
        \label{app-fig.mcq-pipeline-bug.descr}
    \end{subfigure}
    \ \ \ \ \ \ \
    \begin{subfigure}{0.6\textwidth}
        \centering
        \includegraphics[width=0.98\textwidth]{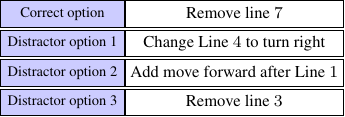}
        \vspace{-1mm}
        \caption{Answer options $\answer + \distractors$}
        \vspace{3mm}
        \label{app-fig.mcq-pipeline-bug.options}
    \end{subfigure}
    \\
    \begin{subfigure}{0.245\textwidth}
        \centering
        \includegraphics[width=0.98\textwidth]{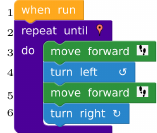}
        \vspace{-1.5mm}
        \caption{Input code \code{}}
        \vspace{1mm}
        \label{app-fig.mcq-pipeline-bug.code}
    \end{subfigure}
    \begin{subfigure}{0.245\textwidth}
        \centering
        \includegraphics[width=0.98\textwidth]{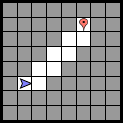}
        \vspace{-4mm}
        \caption{Input grid \grid{}}
        \vspace{1mm}
        \label{app-fig.mcq-pipeline-bug.grid}
    \end{subfigure}
    \begin{subfigure}{0.245\textwidth}
        \centering
        \includegraphics[width=0.98\textwidth]{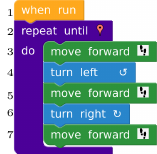}
        \vspace{-4mm}
        \caption{Mutated code $\code_{\text{mut}}$}
        \vspace{1mm}
        \label{app-fig.mcq-pipeline-bug.grid-mut}
    \end{subfigure}
    \begin{subfigure}{0.245\textwidth}
        \centering
        \includegraphics[width=0.98\textwidth]{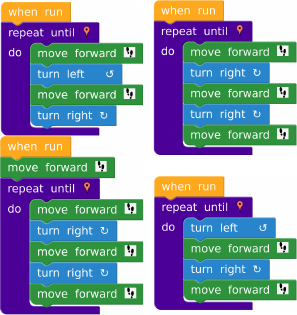}
        \vspace{-4mm}
        \caption{\looseness-1Each option applied to \code{}}
        \vspace{1mm}
        \label{app-fig.mcq-pipeline-bug.valid}
    \end{subfigure}
    \vspace{-2mm}    
    \caption{Illustration of our MCQ task generation methodology instantiated on the bug repair task in \Evaluating{} cognitive level. \textbf{(a)} gives an overview of the pipeline. \textbf{(b,d,e)} show the input components for the pipeline: text description \descr{}, grid \grid{}, and its solution code \code{}. \textbf{(c)} shows the generated correct option \answer{} along with the generated distractors \distractors{}, put together to obtain all four options. \textbf{(f)} shows the mutated code $\code_{\text{mut}}$ obtained according to the selected correct option \answer{}. \textbf{(g)} shows the codes obtained by applying each of the options on \code{}. The MCQ task is valid if the code obtained by applying \answer{} to $\code_{\text{mut}}$ solves \grid{}, and none of the codes obtained by applying \distractors{} to $\code_{\text{mut}}$ solve \grid{}.}
    \label{app-fig.mcq-pipeline-bug}  
\end{figure*}


Next, we exemplify the generation process for a task involving repairing bugs in a given code in Figure~\ref{app-fig.mcq-pipeline-bug}. Similarly to the previous example, we start from a manually written text description for a task \descr{}, a grid \grid{}, and a code \code{}. We first select a code mutation from the pool of feasible mutations. Let us say that the selected mutation is to add a \textcode{move forward} after Line 6. 
We generate the correct answer as the reverse mutation of the previously selected mutation i.e., \answer{} will be to remove Line 7. The next step is to apply the mutation adding \textcode{move forward} after Line 6 on \code{}, thus obtaining $\code_{\text{mut}}$. We then generate distractor options by picking three other feasible mutations that can be applied on $\code_{\text{mut}}$, thus obtaining \distractors{}. Again, after this step, we have all the components of an MCQ task, but the task has to be validated. The validation process for this example involves applying each mutation from the four options on \code{}. A task is valid if \grid{} is solved by the code obtained by applying \answer{} on \code{}, and is not solved by all the other codes obtained by applying \distractors{} on \code{}.

\subsection{Fine-grained Skills Tasks Generation}

We now demonstrate the generation methodology for the fine-grained skills tasks.

\begin{figure*}[!h]
    \begin{subfigure}{0.99\textwidth}
        \centering
        \includegraphics[width=0.98\textwidth]{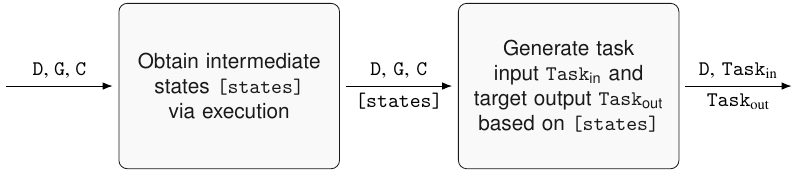}
        \vspace{-1mm}
        \caption{Illustration of the fine-grained skills task generation methodology condition sensing.}
        \vspace{2mm}
        \label{app-fig.basics-pipeline.pipeline}
    \end{subfigure}
    \\
    \begin{subfigure}{0.24\textwidth}
        \centering
        \includegraphics[width=0.98\textwidth]{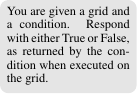}
        \vspace{-4mm}
        \caption{Text description \descr{}}
        \vspace{3mm}
        \label{app-fig.basics-pipeline.descr}
    \end{subfigure}
    \ \
    \begin{subfigure}{0.75\textwidth}
        \centering
        \includegraphics[width=0.98\textwidth]{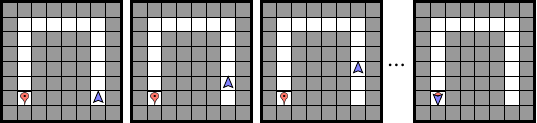}
        \vspace{-1mm}
        \caption{Intermediate states \textcode{[states]}}
        \vspace{3mm}
        \label{app-fig.basics-pipeline.options}
    \end{subfigure}
    \\
    \begin{subfigure}{0.245\textwidth}
        \centering
        \includegraphics[width=0.98\textwidth]{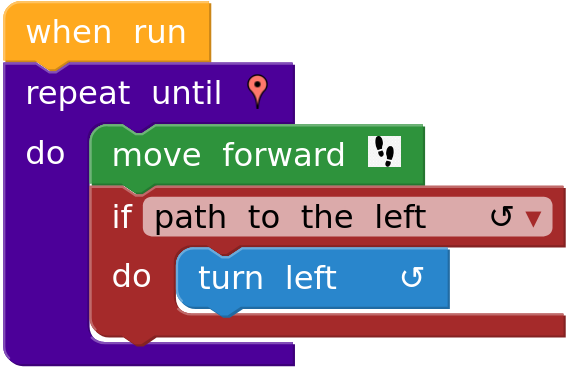}
        \vspace{1.5mm}
        \caption{Input code \code{}}
        \vspace{1mm}
        \label{app-fig.basics-pipeline.code}
    \end{subfigure}
    \begin{subfigure}{0.245\textwidth}
        \centering
        \includegraphics[width=0.98\textwidth]{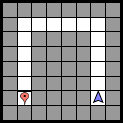}
        \vspace{-4mm}
        \caption{Input grid \grid{}}
        \vspace{1mm}
        \label{app-fig.basics-pipeline.grid}
    \end{subfigure}
    \begin{subfigure}{0.245\textwidth}
        \centering
        \includegraphics[width=0.98\textwidth]{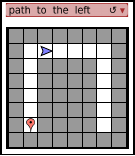}
        \vspace{-4mm}
        \caption{Generated \inp{}}
        \vspace{1mm}
        \label{app-fig.basics-pipeline.grid-mut}
    \end{subfigure}
    \begin{subfigure}{0.245\textwidth}
        \centering
        \includegraphics[width=0.98\textwidth]{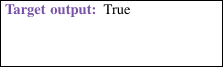}
        \vspace{8mm}
        \caption{Generated \target{}}
        \vspace{1mm}
        \label{app-fig.basics-pipeline.valid}
    \end{subfigure}
    \vspace{-2mm}    
    \caption{Illustration of our fine-grained skills task generation methodology instantiated on the condition sensing basic type of task. \textbf{(a)} gives an overview of the pipeline. \textbf{(b,d,e)} show the input components for the pipeline: text description \descr{}, grid \grid{}, and its solution code \code{}. \textbf{(c)} shows the states generated by executing \code{} on \grid{}. \textbf{(f)} shows the generated content \inp{} which will be part of the task input. \textbf{(g)} shows the generated target output \target{}.}
    \label{app-fig.basics-pipeline}  
\end{figure*}

Figure~\ref{app-fig.basics-pipeline} shows an overview of our fine-grained skills generation methodology applied for a condition sensing task. We start with a code \code{}, a grid \grid{}, and a manually written text description \descr{} which indicates what kind of task should be generated. The first step is to obtain the list of intermediate states \textcode{[states]} by executing \code{} on \grid{}. The intermediate states are grids themselves, with the avatar having various locations and orientations, obtained after each execution of a basic action on the starting grid \grid{}. The next step is to use the information in \textcode{[states]} for generating the input for the task \inp{} and the target output \target{}. In our case, we randomly pick a state from \textcode{[states]}, modify the orientation of the avatar for a richer set of generated situations, and select a random condition from the DSL (i.e., \textcode{path to the left}), thus obtaining \inp{}. We execute the picked condition on the selected state to obtain the target output \target{} (i.e., True). The generation process of other types of tasks is similar.
\clearpage
\section{Additional Data Details}\label{app-sec.data}

\subsection{Hour of Code:Maze Challenge}
The \emph{Hour of Code:Maze Challenge} online programming lesson is publicly accessible at \url{https://studio.code.org/s/hourofcode}. Use of the curriculum is permitted under a Creative Commons BY-NC-SA 4.0 license. We give two more illustrative examples of solution synthesis extracted from \hocType{} in Figure~\ref{app-fig.examples.hoc}.

\begin{figure*}[h!]
    \begin{subfigure}{0.98\textwidth}
        \centering
        \setlength{\fboxsep}{0.05pt}\fbox{\includegraphics[width=0.95\textwidth]{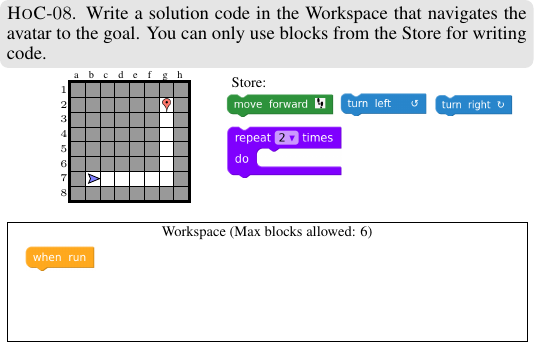}}
        \vspace{-1mm}
        \caption{A \hocType{} task with $\code_{\text{sketch}}=\DSLRepeat{}\textcode{\{\}}$.}
        \vspace{2mm}
        \label{app-fig.examples.hoc.08}
        \end{subfigure}
    \\
    \begin{subfigure}{0.98\textwidth}
        \centering
        \setlength{\fboxsep}{0.05pt}\fbox{\includegraphics[width=0.95\textwidth]{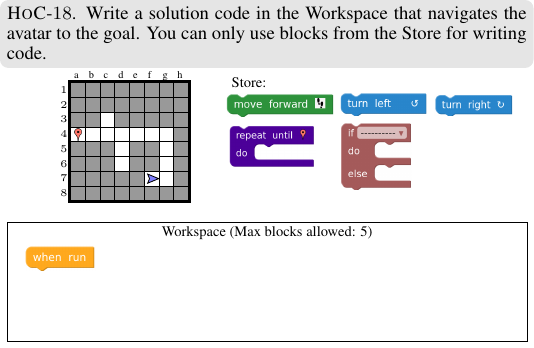}}
        \vspace{-1mm}
        \caption{A \hocType{} task with $\code_{\text{sketch}}=\DSLRepeat{}\textcode{\{\DSLIfElse{}\}}$.}
        \vspace{2mm}
        \label{app-fig.examples.hoc.18}
    \end{subfigure}
    \vspace{-2mm}    
    \caption{Solution synthesis tasks from \hocType{}. Note that the number of times the repeat loop should be executed and the condition of the if statement can be changed.}
    \label{app-fig.examples.hoc}
    \vspace{-5mm}    
\end{figure*}

\clearpage
\subsection{The \aceType{} Test}
This test has been proposed in \cite{DBLP:conf/sigcse/GhoshMS24} and all the MCQ tasks can be found in the Appendix of the respective paper. Next, we give four more illustrative examples for \aceType{}. Figure~\ref{app-fig.examples.ace.06} shows a task corresponding to the \Analyzing{} cognitive level. It involves reasoning about the trace, given a grid and a code. Figures~\ref{app-fig.examples.ace.09}~and~\ref{app-fig.examples.ace.13} show tasks corresponding to the \Evaluating{} cognitive level. The first one involves repairing a buggy code, while the second one involves reasoning about code equivalence without a given grid. Finally, Figure~\ref{app-fig.examples.ace.16} shows a task corresponding to the \Creating{} cognitive level. It involves placing the avatar on the incomplete grid so that the given code solves the modified grid.

\begin{figure*}[h!]
    \begin{subfigure}{0.98\textwidth}
        \centering
        \setlength{\fboxsep}{0.05pt}\fbox{\includegraphics[width=0.95\textwidth]{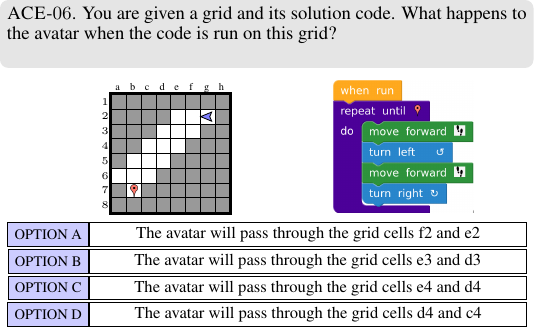}}
        \vspace{-1mm}
        \caption{An \aceType{} task involving tracing.}
        \vspace{2mm}
        \label{app-fig.examples.ace.06}
        \end{subfigure}
    \\
    \begin{subfigure}{0.98\textwidth}
        \centering
        \setlength{\fboxsep}{0.05pt}\fbox{\includegraphics[width=0.95\textwidth]{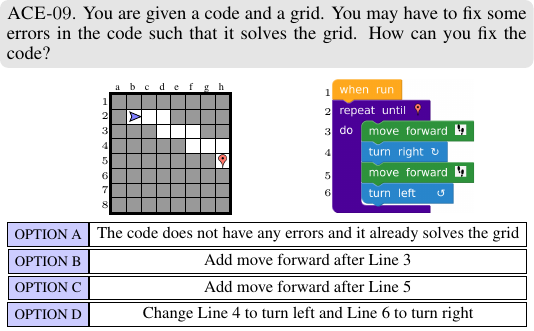}}
        \vspace{-1mm}
        \caption{An \aceType{} task involving bug repair.}
        \vspace{2mm}
        \label{app-fig.examples.ace.09}
    \end{subfigure}
\end{figure*}
\begin{figure*}[h!]\ContinuedFloat
    \begin{subfigure}{0.98\textwidth}
        \centering
        \setlength{\fboxsep}{0.05pt}\fbox{\includegraphics[width=0.95\textwidth]{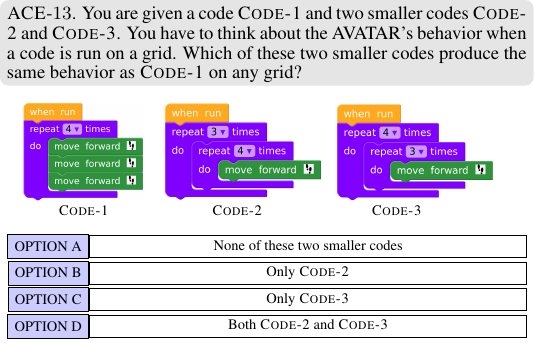}}
        \vspace{-1mm}
        \caption{An \aceType{} task involving code equivalence.}
        \vspace{2mm}
        \label{app-fig.examples.ace.13}
        \end{subfigure}
    \\
    \begin{subfigure}{0.98\textwidth}
        \centering
        \setlength{\fboxsep}{0.05pt}\fbox{\includegraphics[width=0.95\textwidth]{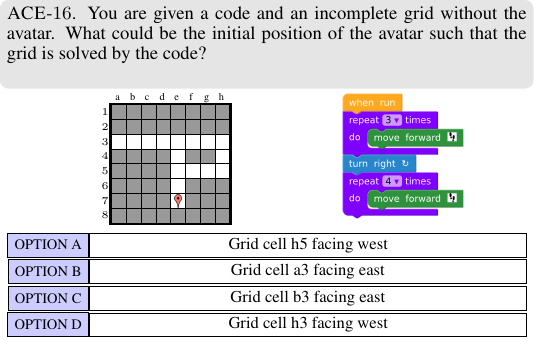}}
        \vspace{-1mm}
        \caption{An \aceType{} task involving avatar design.}
        \vspace{2mm}
        \label{app-fig.examples.ace.16}
    \end{subfigure}
    \vspace{-2mm}    
    \caption{MCQ tasks from \aceType{}.}
    \label{app-fig.examples.ace}
    \vspace{-5mm}    
\end{figure*}

\clearpage
\subsection{The \ctType{}}
This test has been proposed and refined in \cite{gonzalez2015computational,DBLP:journals/chb/Roman-GonzalezP17}, with all the MCQ tasks of the refined version being publicly accessible (in Spanish) at \url{http://goo.gl/IYEKMB}, free of charge, for research purposes. Next, we give two more illustrative examples for \ctType{}. Figure~\ref{app-fig.examples.ct.14} shows a task requiring to pick the solution code for the given grid. Figure~\ref{app-fig.examples.ct.16} shows a task requiring to point out the line that contains an error (i.e., which line should be changed so that the given code becomes a solution for the given grid).

\begin{figure*}[h!]
    \begin{subfigure}{0.98\textwidth}
        \centering
        \setlength{\fboxsep}{0.05pt}\fbox{\includegraphics[width=0.95\textwidth]{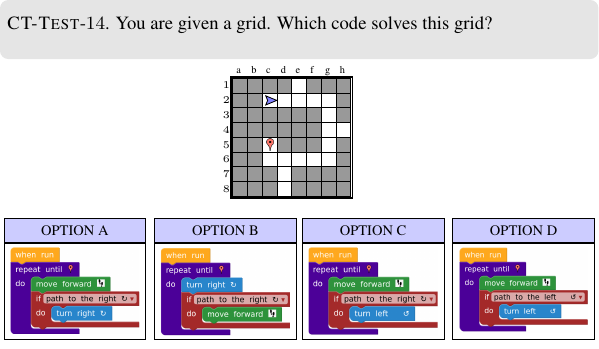}}
        \vspace{-1mm}
        \caption{A \ctType{} task involving picking the solution code.}
        \vspace{2mm}
        \label{app-fig.examples.ct.14}
    \end{subfigure}
    \\
    \begin{subfigure}{0.98\textwidth}
        \centering
        \setlength{\fboxsep}{0.05pt}\fbox{\includegraphics[width=0.95\textwidth]{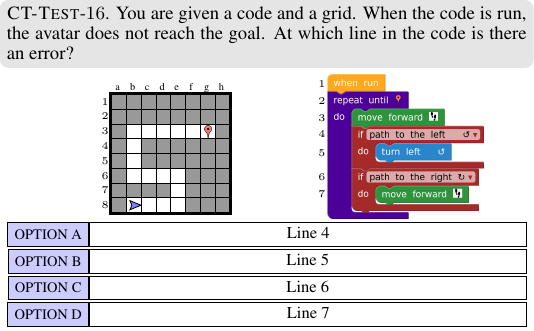}}
        \vspace{-1mm}
        \caption{A \ctType{} task involving pointing out the error.}
        \vspace{2mm}
        \label{app-fig.examples.ct.16}
    \end{subfigure}
    \vspace{-2mm}    
    \caption{MCQ tasks from \ctType{}. \textbf{(a)} shows a task where the correct solution for a given grid needs to be picked. Similar tasks appear in \aceType{}. \textbf{(b)} shows a task involving pointing the line with the error in a given code.}
    \label{app-fig.examples.ct}
    \vspace{-5mm}    
\end{figure*}


\clearpage
\section{Synthetic Evaluation Data}\label{app-sec.synth-data}

Besides our real-world evaluation benchmark, we offer a larger-scale synthetic evaluation segment for \hocType{} and \aceType{}, as has also been considered in the literature \cite{DBLP:conf/iclr/BunelHDSK18,DBLP:conf/iclr/ShinKGBTSS19}. First, we offer details about the composition of our synthetic evaluation segment, then we present results for selected techniques.

\subsection{Generation of the Synthetic Evaluation Segment}

We generate the data using the same generation pipeline. However,  we have the flexibility to vary the nature of synthetic evaluation data and how it differs from its training counterpart. We do this by conditioning the types of generated codes \code{} (i.e., concepts covered, nesting structure). Thus, we obtain four parts of the synthetic evaluation dataset:

\begin{itemize}
    \item \textsc{HoC-synth}: this part corresponds to $758$ solution synthesis tasks generated using the same codes as used for training but with newly generated grids. In particular, the samples encountered during evaluation are similar to samples encountered during training in terms of codes (output) but not in terms of grids (input).
    \item $\textsc{HoC-synth}_\text{hard}$: this part corresponds to $378$ hard solution synthesis tasks that are a subset of $758$ \textsc{HoC-synth} tasks. We selected those tasks for which the solution code requires a condition and are generally more difficult for generative models.
    \item \textsc{HoC-Filtered}: this part corresponds to $758$ solution synthesis tasks. These are also generated using the same generation pipeline as used for generating the training datasets; however, we have ensured that the samples are distinct from training both in terms of codes (output) and grids (inputs).
    \item \textsc{HoC-OoD}: this part corresponds to $100$ solution synthesis tasks, which are “out-of-distribution” (OoD), meaning that the solution structures for these tasks are different and more complex than those used in training datasets. For instance, this set contains tasks that require over three for-loops or a combination of multiple for-loops and while-loops.
    \item \textsc{ACE-synth}: this part corresponds to $922$ MCQ tasks generated using the same codes as used for training but with newly generated grids. Similarly to \textsc{HoC-synth}, the samples encountered during evaluation are similar to samples encountered during training in terms of codes (output) but not in terms of grids (input).
\end{itemize}

\subsection{Results of Selected Models}

\begin{figure*}[h!]
\centering
    \scalebox{0.78}{
        \setlength\tabcolsep{4.5pt}
        \belowrulesep=0pt
        \renewcommand{\arraystretch}{1.2}
        \begin{tabular}{lrrrrr}  
            \toprule
             \multirow[c]{1}{*}[0mm]{{\textbf{Technique}}} & 
             \multicolumn{1}{c}{\textbf{\textsc{HoC-synth}}} &
             \multicolumn{1}{c}{\textbf{$\textsc{HoC-synth}_\text{hard}$}} &
             \multicolumn{1}{c}{\textbf{\textsc{HoC-Filtered}}} &
             \multicolumn{1}{c}{\textbf{\textsc{HoC-OoD}}} &
             \multicolumn{1}{c}{\textbf{\textsc{ACE-synth}}} \\
            \midrule

            $\TechLlamaThreeEight$ &
            $0.1$ $(0.0)$&
            $0.0$ $(0.0)$&
            $0.1$ $(0.0)$&
            $0.0$ $(0.0)$&
            $29.5$ $(0.0)$\\

            $\TechGPTFourOCombined$ &
            $12.8$ $(0.0)$&
            $10.3$ $(0.0)$&
            $20.8$ $(0.0)$&
            $17.0$ $(0.0)$&
            $47.6$ $(1.0)$\\

            $\TechHierFinetuneThree$&
            $37.5$ $(1.0)$&
            $8.6$ $(0.0)$&
            $34.4$ $(2.0)$&
            $18.0$ $(3.0)$&
            $77.5$ $(0.0)$\\
            
            \bottomrule   
        \end{tabular}
        }
    \vspace{-1mm}
    \caption{Results of selected models on different parts of the synthetic evaluation segment.}
    \label{app-fig.synth-data.results}
    \vspace{-3mm}
\end{figure*}

We evaluate each of the following selected techniques over three seeds: \TechLlamaThreeEight{}, \TechGPTFourOCombined{}, and \TechHierFinetuneThree{}. Figure~\ref{app-sec.synth-data} shows the results of the selected techniuqes, averaged as mean (stderr).\TechLlamaThreeEight{} struggles consistently across all parts of the synthetic evaluation dataset, showing poor performance in both simple and challenging tasks. \TechGPTFourOCombined{} struggles more with synthetic data compared to real-world tasks (results shown in Figure~\ref{fig.experiments.results_table}). \TechHierFinetuneThree{} generally outperforms both models on synthetic data, showing strong results across most tasks. It also demonstrates good generalization capabilities, outperforming \TechGPTFourOCombined{} on the \textsc{HoC-Filtered} set and achieving comparable performance on the out-of-distribution tasks (\textsc{HoC-OoD}).
\clearpage
\section{Additional Reasoning Examples}\label{app-sec.reasoning}

\looseness-1We provide reasoning examples done by \TechHierFinetuneThree{}, \TechHocFinetuneNoExpThree{}, and \TechGPTFourOCombined{}. First, we show how explanations help \TechLlamaCT{} reason better spatially. For the solution synthesis task in \iftoggle{MainSuppContent}{Figure~\ref{fig.illustration.hoc}}{{Figure~1b}}, \TechHierFinetuneThree{} provides an almost correct explanation (shown in Figure~\ref{fig.reasoning.explanations.hoc_ours}), helping the model solve the task by correctly understanding the avatar's orientation. In contrast, \TechHocFinetuneNoExpThree{} (see Figure~\ref{fig.reasoning.explanations.no_exp}) struggles with distinguishing left from right, failing to provide a correct solution despite understanding the task's general structure. Next, we provide reasoning examples for \TechGPTFourOCombined{} for the tasks in \iftoggle{MainSuppContent}{Figures~\ref{fig.illustration.hoc}~and~\ref{fig.illustration.ace}}{Figures~1b~and~1c}. The parts where the reasoning is wrong are highlighted in red. We can notice that for \hocType{}, the model comes up with a general code that may be able to solve a large variety of grids, but it doesn't respect \limSize{} and \codeStore{} from \iftoggle{MainSuppContent}{Figure~\ref{fig.illustration.hoc}}{Figure~1b}. Additionally, the reasoning involving tracing and the sequence of basic actions is flawed, as the model doesn't take all walls into account. The model makes a very convincing argument for the \aceType{} example, yet it reaches a wrong solution by a partially flawed reasoning process. It places walls in a manner that would block the avatar from reaching the goal in \iftoggle{MainSuppContent}{Figure~\ref{fig.illustration.ace}}{Figure~1c}.

\begin{figure*}[h!]
        \hspace{-10mm}
        \begin{subfigure}{0.79\linewidth}
        {
        \centering
        \scalebox{0.75}{
             \begin{tabular}{|p{1\linewidth}|}
                \hline
                \multicolumn{1}{|p{1\linewidth}|}
                {                                   
                    {\small \input{appendix_figs/reasoning/llamact/content/hoc_ours}}
                }\\[-1.5ex]
                \hline
            \end{tabular}
            }
            \vspace{-1.25mm}
            \caption{Answer with an explanation.}				
            \label{fig.reasoning.explanations.hoc_ours}
        }
        \end{subfigure}
        \hspace{-10mm}
        \begin{subfigure}{0.3\linewidth}
        {
        \centering
        \scalebox{0.75}{
             \begin{tabular}{|p{1\linewidth}|}
                \hline
                \multicolumn{1}{|p{1\linewidth}|}
                {                                   
                    {\small \input{appendix_figs/reasoning/llamact/content/hoc_no_exp}}
                }\\[-1ex]
                \hline
            \end{tabular}
            }
            \vspace{-1.25mm}
            \caption{Answer with no explanation.}				
            \label{fig.reasoning.explanations.no_exp}
        }
        \end{subfigure}
    \vspace{-2.25mm}
    \caption{
        Illustrative examples showing the reasoning processes of \TechHierFinetuneThree{} and \TechHocFinetuneNoExpThree{} for \hocType{} task in Figure~\ref{fig.illustration.hoc}, with errors highlighted in red.
    }
    \label{fig.reasoning.explanations}
    \vspace{-2mm}
\end{figure*}

\begin{figure*}[h!]
\centering
    \begin{subfigure}{0.98\linewidth}
    {
         \begin{tabular}{!{\vrule}p{1\linewidth}!{\vrule}}
            \hline
            \multicolumn{1}{!{\vrule}p{1\linewidth}!{\vrule}}
            {                                   
                {\small \input{appendix_figs/reasoning/gpt/content/hoc_gpt4ovt}}
            }\\
            \hline
        \end{tabular}
        \vspace{-2mm}
        \caption{Reasoning for \hocType{}.}				
        \label{app-fig.reasoning.hoc_gpt}
    }
    \end{subfigure}
    \\
    \begin{subfigure}{0.98\linewidth}
    {
         \begin{tabular}{!{\vrule}p{1\linewidth}!{\vrule}}
            \hline
            \multicolumn{1}{!{\vrule}p{1\linewidth}!{\vrule}}
            {                                   
                {\small \input{appendix_figs/reasoning/gpt/content/ace_gpt4ovt}}
            }\\
            \hline
        \end{tabular}
        \vspace{-2mm}
        \caption{Reasoning for \aceType{}.}				
        \label{app-fig.reasoning.ace_gpt}
    }
    \end{subfigure}
    \vspace{-1mm}
    \caption{
        Examples of reasoning done by \TechGPTFourOCombined{}.
    }
    \label{app-fig.reasoning}
    \vspace{-6mm}
\end{figure*}
\clearpage
\section{Prompt Templates}\label{app-sec.prompts}

In this section, we give details about the prompts we used for fine-tuning \TechLlamaCT{} and for inference with all models.

\subsection{Domain background for \hocType{} and \aceType{}}
We start with the prompts that offer background knowledge and familiarize the models with our representations of the grids and codes for the \hocType{} and \aceType{} tests. Figure~\ref{app-fig.prompts.background.text} presents the prompt used to provide background information during fine-tuning. The same prompt was used during inference for models receiving grids under text representation. Figure~\ref{app-fig.prompts.background.vision} shows the prompt used during inference with models for which we input grids under visual representation. Finally, Figure~\ref{app-fig.prompts.background.visionandtext} shows the prompt used during inference for models taking as input grids under both text and visual representation.

\begin{figure*}[ht]
\centering
	\begin{subfigure}[b]{1\textwidth}
    	\centering
        \scalebox{0.73}{
            \setlength\tabcolsep{2.5pt}
            \renewcommand{\arraystretch}{1.5}
            \begin{tabular}{|p{1.3\linewidth}|}     
                \hline
                \multicolumn{1}{|c|}{\textcolor{RoyalPurple}{\textbf{Background information and representation for text}}} \\
                \input{appendix_figs/prompts/content/background_text}
                \\
                \hline
                
            \end{tabular}
            }
        \vspace{2mm}
        \caption{Prompt used as background information for models employing text modality only.}
        \label{app-fig.prompts.background.text}
    \end{subfigure}
\end{figure*}
\begin{figure*}[ht]\ContinuedFloat
\centering
	\begin{subfigure}[b]{1\textwidth}
    	\centering
        \scalebox{0.73}{
            \setlength\tabcolsep{2.5pt}
            \renewcommand{\arraystretch}{1.5}
            \begin{tabular}{|p{1.3\linewidth}|}     
                \hline
                \multicolumn{1}{|c|}{\textcolor{RoyalPurple}{\textbf{Background information and representation for vision}}} \\
                \input{appendix_figs/prompts/content/background_vision}
                \\
                \hline
                
            \end{tabular}
            }
        \vspace{2mm}
        \caption{Prompt used as background information for models employing vision modality only.}
        \label{app-fig.prompts.background.vision}
    \end{subfigure}
    \\~\\~\\
	\begin{subfigure}[b]{1\textwidth}
    	\centering
        \scalebox{0.73}{
            \setlength\tabcolsep{2.5pt}
            \renewcommand{\arraystretch}{1.25}
            \begin{tabular}{|p{1.35\linewidth}|}     
                \hline
                \multicolumn{1}{|c|}{\textcolor{RoyalPurple}{\textbf{Background information and representation for vision+text}}} \\
                \input{appendix_figs/prompts/content/background_vision_text}
                \\
                \hline
                
            \end{tabular}
            }
        \vspace{2mm}
        \caption{Prompt used as background information for models employing both vision and text modalities.}
        \label{app-fig.prompts.background.visionandtext}
    \end{subfigure}
    \vspace{5mm}
    \caption{Prompts used for offering background and representation information for \hocType{} and \aceType{}.}
    \label{app-fig.prompts.background}
\end{figure*}%


\clearpage
\subsection{Domain background for \ctType{}}
Similarly, we present the prompts offering background and representation information during inference on \ctType{}. These include some additional information involving Karel tasks and another type of tasks containing colored cells. Figure~\ref{app-fig.prompts.background-ct.text} shows the prompt used for models receiving grids under text representation. Figure~\ref{app-fig.prompts.background-ct.vision} shows the prompt used for models receiving grids under visual representation only. Finally, Figure~\ref{app-fig.prompts.background-ct.visionandtext} shows the prompt used during inference for models taking as input grids under both text and visual representation.

\begin{figure*}[ht]
\centering
	\begin{subfigure}[b]{1\textwidth}
    	\centering
        \scalebox{0.73}{
            \setlength\tabcolsep{2.5pt}
            \renewcommand{\arraystretch}{1.5}
            \begin{tabular}{|p{1.3\linewidth}|}        
                \hline
                \multicolumn{1}{|c|}{\textcolor{RoyalPurple}{\textbf{Background information and representation for text}}} \\
                \input{appendix_figs/prompts/content/background_text_ct}
                \\
                \hline
                
            \end{tabular}
            }
        \vspace{2mm}
        \caption{Prompt used as background information for models employing text modality only.}
        \label{app-fig.prompts.background-ct.text}
    \end{subfigure}
\end{figure*}
\begin{figure*}[ht]\ContinuedFloat
\centering
	\begin{subfigure}[b]{1\textwidth}
    	\centering
        \scalebox{0.73}{
            \setlength\tabcolsep{2.5pt}
            \renewcommand{\arraystretch}{1.5}
            \begin{tabular}{|p{1.3\linewidth}|}        
                \hline
                \multicolumn{1}{|c|}{\textcolor{RoyalPurple}{\textbf{Background information and representation for vision}}} \\
                \input{appendix_figs/prompts/content/background_vision_ct}
                \\
                \hline
                
            \end{tabular}
            }
        \vspace{2mm}
        \caption{Prompt used as background information for models employing vision modality only.}
        \label{app-fig.prompts.background-ct.vision}
    \end{subfigure}
    \\~\\~\\
	\begin{subfigure}[b]{1\textwidth}
    	\centering
        \scalebox{0.73}{
            \setlength\tabcolsep{2.5pt}
            \renewcommand{\arraystretch}{1.5}
            \begin{tabular}{|p{1.3\linewidth}|}     
                \hline
                \multicolumn{1}{|c|}{\textcolor{RoyalPurple}{\textbf{Background information and representation for vision+text}}} \\
                \input{appendix_figs/prompts/content/background_vision_text_ct}
                \\
                \hline
                
            \end{tabular}
            }
        \vspace{2mm}
        \caption{Prompt used as background information for models employing both vision and text modalities.}
        \label{app-fig.prompts.background-ct.visionandtext}
    \end{subfigure}
    \vspace{5mm}
    \caption{Prompts used for offering background and representation information for \ctType{}.}
    \label{app-fig.prompts.background-ct}
\end{figure*}%


\clearpage
\subsection{Instructions}
We continue with the prompts used to instruct the models about their task. We use the prompt in Figure~\ref{app-fig.prompts.instruction.solsyn} for inference on \hocType{}, and the prompt in Figure~\ref{app-fig.prompts.instruction.mcq} for inference on \aceType{} and \ctType{}. All the prompts in Figure~\ref{app-fig.prompts.instruction} were used for fine-tuning.

\begin{figure*}[h!]
\centering
	\begin{subfigure}[b]{1\textwidth}
    	\centering
        \scalebox{0.73}{
            \setlength\tabcolsep{2.5pt}
            \renewcommand{\arraystretch}{1.5}
            \begin{tabular}{|p{1.3\linewidth}|}     
                \hline
                \multicolumn{1}{|c|}{\textcolor{RoyalPurple}{\textbf{Solution synthesis prompt}}} \\
                \input{appendix_figs/prompts/content/solution_synthesis}
                \\
                \hline
            \end{tabular}
            }
        \caption{Prompt used for solution synthesis.}
        \label{app-fig.prompts.instruction.solsyn}
    \end{subfigure}
    \\~\\
    \begin{subfigure}[b]{1\textwidth}
    	\centering
        \scalebox{0.73}{
            \setlength\tabcolsep{2.5pt}
            \renewcommand{\arraystretch}{1.5}
            \begin{tabular}{|p{1.3\linewidth}|}     
                \hline
                \multicolumn{1}{|c|}{\textcolor{RoyalPurple}{\textbf{Multi-choice question prompt}}} \\
                \input{appendix_figs/prompts/content/mcq}
                \\
                \hline
            \end{tabular}
            }
        \caption{Prompt used for asking multi-choice questions.}
        \label{app-fig.prompts.instruction.mcq}
    \end{subfigure}
    \\~\\
	\begin{subfigure}[b]{1\textwidth}
    	\centering
        \scalebox{0.73}{
            \setlength\tabcolsep{2.5pt}
            \renewcommand{\arraystretch}{1.5}
            \begin{tabular}{|p{1.3\linewidth}|}     
                \hline
                \multicolumn{1}{|c|}{\textcolor{RoyalPurple}{\textbf{Locate avatar prompt}}} \\
                \input{appendix_figs/prompts/content/locate_avatar}
                \\
                \hline
            \end{tabular}
            }
        \caption{Prompt used for locating the avatar in the basics fine-grained skills.}
        \label{app-fig.prompts.instruction.locate-avatar}
    \end{subfigure}
    \\~\\
	\begin{subfigure}[b]{1\textwidth}
    	\centering
        \scalebox{0.73}{
            \setlength\tabcolsep{2.5pt}
            \renewcommand{\arraystretch}{1.5}
            \begin{tabular}{|p{1.3\linewidth}|}     
                \hline
                \multicolumn{1}{|c|}{\textcolor{RoyalPurple}{\textbf{Locate goal prompt}}} \\
                \input{appendix_figs/prompts/content/locate_goal}
                \\
                \hline
            \end{tabular}
            }
        \caption{Prompt used for locating the goal in the basics fine-grained skills.}
        \label{app-fig.prompts.instruction.locate-goal}
    \end{subfigure}
    \\~\\
	\begin{subfigure}[b]{1\textwidth}
    	\centering
        \scalebox{0.73}{
            \setlength\tabcolsep{2.5pt}
            \renewcommand{\arraystretch}{1.5}
            \begin{tabular}{|p{1.3\linewidth}|}     
                \hline
                \multicolumn{1}{|c|}{\textcolor{RoyalPurple}{\textbf{Apply action prompt}}} \\
                \input{appendix_figs/prompts/content/apply_action}
                \\
                \hline
            \end{tabular}
            }
        \caption{Prompt used for applying an action in the basics fine-grained skills.}
        \label{app-fig.prompts.instruction.act}
        \vspace{-5mm}
    \end{subfigure}
\end{figure*}%
\begin{figure*}[h!]\ContinuedFloat
\centering
	\begin{subfigure}[b]{1\textwidth}
    	\centering
        \scalebox{0.73}{
            \setlength\tabcolsep{2.5pt}
            \renewcommand{\arraystretch}{1.5}
            \begin{tabular}{|p{1.3\linewidth}|}     
                \hline
                \multicolumn{1}{|c|}{\textcolor{RoyalPurple}{\textbf{Sense condition prompt}}} \\
                \input{appendix_figs/prompts/content/sense_condition}
                \\
                \hline
            \end{tabular}
            }
        \caption{Prompt used for sensing a condition in the basics fine-grained skills.}
        \label{app-fig.prompts.instruction.sense}
    \end{subfigure}
    \\~\\
    \begin{subfigure}[b]{1\textwidth}
    	\centering
        \scalebox{0.73}{
            \setlength\tabcolsep{2.5pt}
            \renewcommand{\arraystretch}{1.5}
            \begin{tabular}{|p{1.3\linewidth}|}     
                \hline
                \multicolumn{1}{|c|}{\textcolor{RoyalPurple}{\textbf{Trace sequence prompt}}} \\
                \input{appendix_figs/prompts/content/trace_seq}
                \\
                \hline
            \end{tabular}
            }
        \caption{Prompt used for tracing a sequence of basic actions in the tracing fine-grained skills.}
        \label{app-fig.prompts.instruction.trace-seq}
    \end{subfigure}
    \\~\\
    \begin{subfigure}[b]{1\textwidth}
    	\centering
        \scalebox{0.73}{
            \setlength\tabcolsep{2.5pt}
            \renewcommand{\arraystretch}{1.5}
            \begin{tabular}{|p{1.3\linewidth}|}     
                \hline
                \multicolumn{1}{|c|}{\textcolor{RoyalPurple}{\textbf{Trace code prompt}}} \\
                \input{appendix_figs/prompts/content/trace_code}
                \\
                \hline
            \end{tabular}
            }
        \caption{Prompt used for tracing a code in the tracing fine-grained skills.}
        \label{app-fig.prompts.instruction.trace-code}
    \end{subfigure}
    \\~\\
    \begin{subfigure}[b]{1\textwidth}
    	\centering
        \scalebox{0.73}{
            \setlength\tabcolsep{2.5pt}
            \renewcommand{\arraystretch}{1.5}
            \begin{tabular}{|p{1.3\linewidth}|}     
                \hline
                \multicolumn{1}{|c|}{\textcolor{RoyalPurple}{\textbf{Place avatar prompt}}} \\
                \input{appendix_figs/prompts/content/place_avatar}
                \\
                \hline
            \end{tabular}
            }
        \caption{Prompt used for placing the avatar in the grid synthesis fine-grained skills.}
        \label{app-fig.prompts.instruction.place-avatar}
    \end{subfigure}
\end{figure*}
\begin{figure*}[h!]\ContinuedFloat
\centering
	\begin{subfigure}[b]{1\textwidth}
    	\centering
        \scalebox{0.73}{
            \setlength\tabcolsep{2.5pt}
            \renewcommand{\arraystretch}{1.5}
            \begin{tabular}{|p{1.3\linewidth}|}     
                \hline
                \multicolumn{1}{|c|}{\textcolor{RoyalPurple}{\textbf{Place goal prompt}}} \\
                \input{appendix_figs/prompts/content/place_goal}
                \\
                \hline
            \end{tabular}
            }
        \caption{Prompt used for placing the goal in the grid synthesis fine-grained skills.}
        \label{app-fig.prompts.instruction.place-goal}
    \end{subfigure}
    \\~\\
    \begin{subfigure}[b]{1\textwidth}
    	\centering
        \scalebox{0.73}{
            \setlength\tabcolsep{2.5pt}
            \renewcommand{\arraystretch}{1.5}
            \begin{tabular}{|p{1.3\linewidth}|}     
                \hline
                \multicolumn{1}{|c|}{\textcolor{RoyalPurple}{\textbf{Place avatar+goal prompt}}} \\
                \input{appendix_figs/prompts/content/place_avatar_goal}
                \\
                \hline
            \end{tabular}
            }
        \caption{Prompt used for placing both the avatar and the goal in the grid synthesis fine-grained skills.}
        \label{app-fig.prompts.instruction.place-avatargoal}
    \end{subfigure}
    \\~\\
    \begin{subfigure}[b]{1\textwidth}
    	\centering
        \scalebox{0.73}{
            \setlength\tabcolsep{2.5pt}
            \renewcommand{\arraystretch}{1.5}
            \begin{tabular}{|p{1.3\linewidth}|}     
                \hline
                \multicolumn{1}{|c|}{\textcolor{RoyalPurple}{\textbf{Place walls prompt}}} \\
                \input{appendix_figs/prompts/content/place_walls}
                \\
                \hline
            \end{tabular}
            }
        \caption{Prompt used for placing walls in the grid synthesis fine-grained skills.}
        \label{app-fig.prompts.instruction.place-walls}
    \end{subfigure}
    \\~\\
    \begin{subfigure}[b]{1\textwidth}
    	\centering
        \scalebox{0.73}{
            \setlength\tabcolsep{2.5pt}
            \renewcommand{\arraystretch}{1.5}
            \begin{tabular}{|p{1.3\linewidth}|}     
                \hline
                \multicolumn{1}{|c|}{\textcolor{RoyalPurple}{\textbf{Design all prompt}}} \\
                \input{appendix_figs/prompts/content/design_all}
                \\
                \hline
            \end{tabular}
            }
        \caption{Prompt used for designing the full grid in the grid synthesis fine-grained skills.}
        \label{app-fig.prompts.instruction.fullgrid}
    \end{subfigure}
    \caption{Prompts with placeholders for input data used as instructions for each type of task, without correct reasoning provided at inference. Some of the prompts mention that the model should not solve the task. This is required for making the model reason about the given instruction, and not simply write a code that would solve the grid.}
    \label{app-fig.prompts.instruction}
\end{figure*}

\clearpage
\subsection{Instructions with correct reasoning at inference}
Figure~\ref{app-fig.prompts.instruction-emu} shows the prompts replacing the ones in Figures~\ref{app-fig.prompts.instruction.solsyn}~and~\ref{app-fig.prompts.instruction.mcq} for fine-tuning and inference with \TechHierEmuFinetuneEmuInfThree{}, that receive the correct reasoning during inference.

\begin{figure*}[h!]
\centering
	\begin{subfigure}[b]{1\textwidth}
    	\centering
        \scalebox{0.73}{
            \setlength\tabcolsep{2.5pt}
            \renewcommand{\arraystretch}{1.5}
            \begin{tabular}{|p{1.3\linewidth}|}     
                \hline
                \multicolumn{1}{|c|}{\textcolor{RoyalPurple}{\textbf{Solution synthesis with correct reasoning prompt}}} \\
                \input{appendix_figs/prompts/content/solution_synthesis_emu}
                \\
                \hline
            \end{tabular}
            }
        \caption{Prompt used for solution synthesis.}
        \label{app-fig.prompts.instruction-emu.solsyn}
    \end{subfigure}
    \\~\\
	\begin{subfigure}[b]{1\textwidth}
    	\centering
        \scalebox{0.73}{
            \setlength\tabcolsep{2.5pt}
            \renewcommand{\arraystretch}{1.5}
            \begin{tabular}{|p{1.3\linewidth}|}     
                \hline
                \multicolumn{1}{|c|}{\textcolor{RoyalPurple}{\textbf{Multi-choice question with correct reasoning prompt}}} \\
                \input{appendix_figs/prompts/content/mcq_emu}
                \\
                \hline
            \end{tabular}
            }
        \caption{Prompt used for multi-choice questions.}
        \label{app-fig.prompts.instruction-emu.mcq}
    \end{subfigure}
    \caption{Prompts with placeholders for input data used as instructions for the two target tasks, with correct reasoning provided at inference.}
    \label{app-fig.prompts.instruction-emu}
\end{figure*}
}
}
{
}


\end{document}